\documentclass[lettersize,journal]{IEEEtran}
\usepackage{amsmath,amsfonts}
\usepackage{algorithmic}
\usepackage{algorithm}
\usepackage{array}
\usepackage[caption=false,font=normalsize,labelfont=sf,textfont=sf]{subfig}
\usepackage{textcomp}
\usepackage{stfloats}
\usepackage{url}
\usepackage{verbatim}
\usepackage{graphicx}
\usepackage{cite}
\newcommand{\pub}[1]{\color{gray}{\tiny{[{#1}]}}}
\usepackage{newfloat}
\usepackage{listings}
\usepackage{multirow}
\usepackage{color}
\usepackage{bm}
\usepackage{bbding}
\usepackage{pifont}
\usepackage{makecell}
\usepackage{colortbl}
\usepackage[table]{xcolor}
\usepackage{microtype}
\usepackage[colorlinks,linkcolor=red,anchorcolor=blue,citecolor=green]{hyperref}
\usepackage{booktabs}
\usepackage{wrapfig}
\newfloat{figtab}{htb}{fgtb}
\usepackage{subfiles}
\newcommand{\cmark}{\ding{51}}%
\definecolor{cred}{HTML}{FF6B6B}
\definecolor{cyellow}{HTML}{FEC260}
\definecolor{cgreen}{HTML}{70AD47}
\definecolor{cblue}{HTML}{4D96FF}
\definecolor{cpurple}{HTML}{2A0944}
\definecolor{ggray}{RGB}{127,127,127}
\definecolor{aliceblue}{rgb}{0.94, 0.97, 1.0}
\definecolor{bblue}{RGB}{0,30,95}
\definecolor{rred}{RGB}{190,0,0}
\definecolor{mygray}{gray}{.9}
\definecolor{ggray}{RGB}{127,127,127}
\definecolor{sblue}{RGB}{0,173,206}
\definecolor{ppink}{RGB}{240,46,142}

\def\eg{\emph{e.g.}} 
\def\ie{\emph{i.e.}} 
\def\cf{\emph{c.f.}} 
\def\etc{\emph{etc.}} 
\usepackage{colortbl} 
\definecolor{codegreen}{RGB}{79,126,127}
\definecolor{codedefine}{RGB}{153,54,159}
\definecolor{codefunc}{RGB}{73,122,234}
\definecolor{codecall}{RGB}{73,122,234}
\definecolor{codepro}{RGB}{212,96,80}
\definecolor{codedim}{RGB}{89,152,195}
\definecolor{codeorg}{RGB}{109,162,200}

\makeatletter
\newcommand{\thickhline}{%
    \noalign {\ifnum 0=`}\fi \hrule height 1pt
    \futurelet \reserved@a \@xhline
}

\begin{document}

\title{Locality-aware Cross-modal Correspondence Learning for Dense Audio-Visual Events Detection}

\author{IEEE Publication Technology,~\IEEEmembership{Staff,~IEEE,}
\thanks{This paper was produced by the IEEE Publication Technology Group. They are in Piscataway, NJ.}
\thanks{Manuscript received April 19, 2021; revised August 16, 2021.}}


\author{Ling~Xing, Hongyu~Qu, Rui~Yan, Xiangbo~Shu, and Jinhui~Tang
\thanks{\textit{L. Xing, H. Qu, R. Yan, X. Shu, and J. Tang are with the School of Computer Science and Engineering, Nanjing University of Science and Technology, Nanjing 210094, China. E-mail: \{lingxing, quhongyu, ruiyan, shuxb, jinhuitang\}@njust.edu.cn.}}
}

\maketitle
\begin{abstract}
Dense-localization Audio-Visual Events (DAVE) aims to identify time boundaries and corresponding categories for events that are both audible and visible in a long video, where events may co-occur and exhibit varying durations.
However, complex audio-visual scenes often involve asynchronization between modalities, making accurate localization challenging. 
Existing DAVE solutions extract audio and visual features through unimodal encoders, and fuse them via dense cross-modal interaction. 
However, independent unimodal encoding \textit{struggles to emphasize shared semantics between modalities} without cross-modal guidance, while dense cross-modal attention may \textit{over-attend to semantically unrelated audio-visual features}.
To address these problems, we present \textbf{\textsc{LoCo}}, a \underline{\textbf{Lo}}cality-aware cross-modal \underline{\textbf{Co}}rrespondence learning framework for DAVE. 
\textsc{LoCo} leverages the local temporal continuity of audio-visual events as important guidance to filter irrelevant cross-modal signals and enhance cross-modal alignment throughout both unimodal and cross-modal encoding stages.
\textbf{i)} Specifically, \textsc{LoCo} applies Local Correspondence Feature (LCF) Modulation to enforce unimodal encoders to focus on modality-shared semantics by modulating agreement between audio and visual features based on local cross-modal coherence. 
\textbf{ii)} To better aggregate cross-modal relevant features, we further customize Local Adaptive Cross-modal  (LAC) Interaction, which dynamically adjusts attention regions in a data-driven manner. 
This adaptive mechanism focuses attention on local event boundaries and accommodates varying event durations.
By incorporating LCF and LAC, \textsc{LoCo} provides solid performance gains and outperforms existing DAVE methods. 
The source code will be released. 

\end{abstract}
\begin{IEEEkeywords}
Audio-visual events localization, Local cross-modal
coherence, Cross-modal correspondence learning
\end{IEEEkeywords}

\begin{figure}[t]
    \centering
    \includegraphics[width=0.49\textwidth]{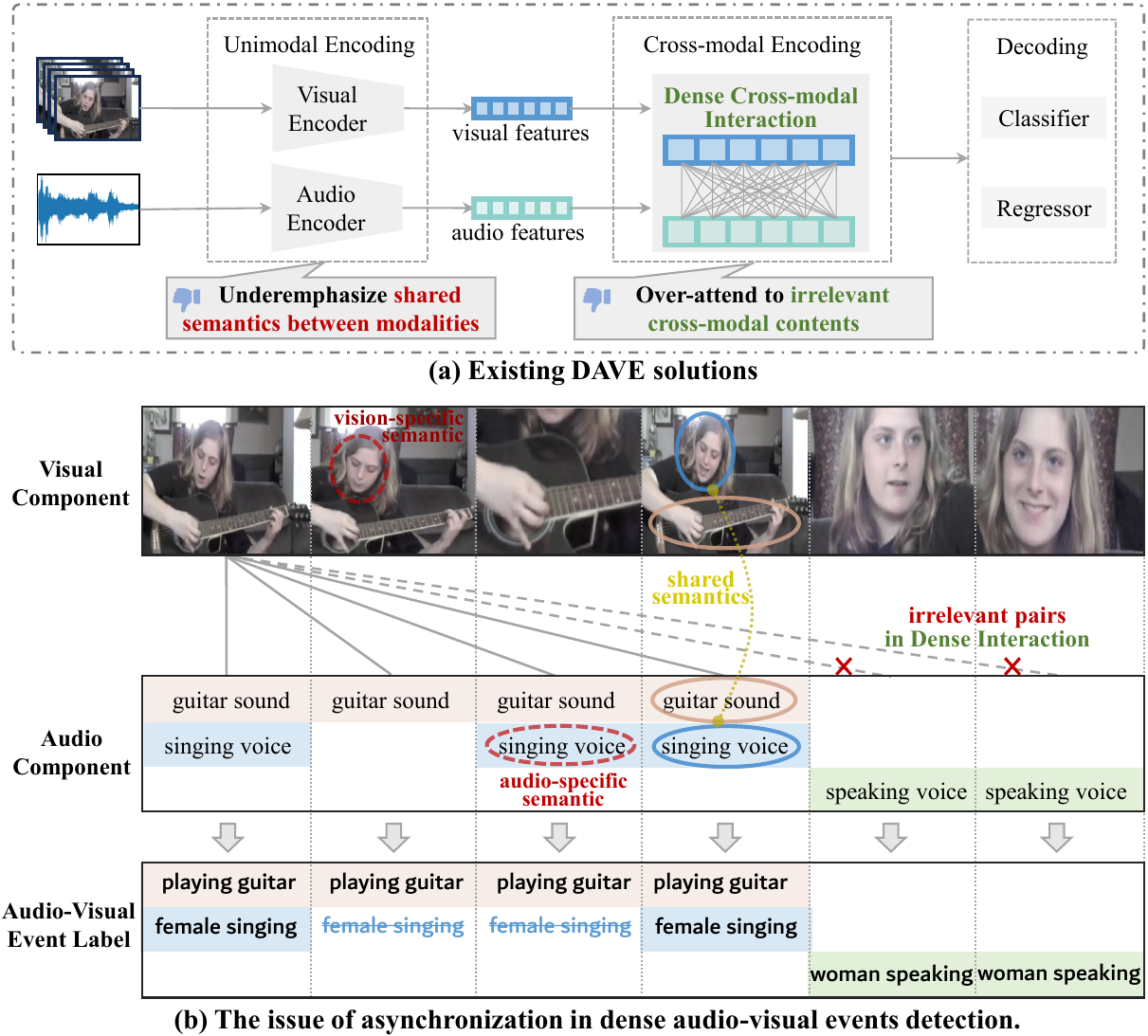} 
   \caption{Existing DAVE methods typically extract audio
and visual features using separate unimodal encoders (\ie, unimodal encoding stage), and fuse them through dense cross-attention interaction. \textbf{Such solutions suffer from two key issues.} \textbf{i)} Independent unimodal encoding underemphasizes shared semantics between audio and visual signals in the absence of cross-modal mutual guidance, hindering the ability of the model to suppress modality-specific noise (\eg, the red dashed circles in (b)). 
   \textbf{ii) }Dense cross-attention interaction over-attends to irrelevant cross-modal contents (\eg, the gray dashed lines in (b)), introducing semantic confusion.
   }
    \label{motivation}
\end{figure}




\section{Introduction}
In real-world scenarios, events often span multiple modalities that are inherently correlated~\cite{chatterjee2022learning, zhang2022morphmlp, zhang2023show, qu2024mvp, huang2024appearance,chen2022comprehensive,he2022multi,luo2023multi,cao2023unsupervised,mou2023compressed}. 
To enhance the perception of the world through audio and visual signals, Audio-Visual Event Localization (AVEL)~\cite{AVE} has been widely explored, which seeks to identify a single audio-visual event (\ie, both audible and visible) within a short, trimmed video (\eg, 10s).
While AVEL has achieved significant progress, its simplified setting falls short of capturing the complexity of real-world scenarios, where multiple events co-occur and evolve over longer temporal spans.

In this paper, we explore a more practical task, Dense-localizing Audio-Visual Events (DAVE)~\cite{geng2023dense}, which aims at recognizing and localizing multiple audio-visual events in a long untrimmed video (\eg, 60s). 
DAVE allows events to overlap in time and exhibit varying durations. 
Although both AVEL and DAVE involve understanding audio-visual events, they are \textbf{\textit{fundamentally different}} in terms of task formulation. 
AVEL is defined as a \textbf{classification task} at the segment level for trimmed video~\cite{AVE,xu2020cross,xuan2020cross,ge2023learning}, whereas DAVE requires \textbf{frame-level regression} to accurately localize events within long untrimmed video~\cite{geng2023dense,geng2024uniav}. 

Complex audio-visual scenes often involve asynchronization between audio and visual cues, making it challenging to capture accurate cross-modal correspondence.
Existing DAVE solutions$_{\!}$~\cite{geng2023dense,geng2024uniav} typically extract audio and visual features with separate unimodal encoders, and fuse them via dense cross-attention interaction, as illustrated in Fig.$_{\!}$~\ref{motivation} $_{\!}$(a).
Though straightforward, they have two key limitations: 
\textbf{First}, independent unimodal encoding \textit{ignores the shared semantics} between audio and visual streams, which may overemphasize modality-specific semantics without cross-modal guidance.
This issue is more pronounced in complex scenes, where the audio and visual tracks may mismatch, \eg, the audio contains singing voices, but no visible singer (the red dashed circles in Fig.$_{\!}$~\ref{motivation} $_{\!}$(b)). 
The lack of early cross-modal interaction limits the ability of the model to suppress modality-specific noise.
\textbf{Second}, dense cross-modal attention interaction attends to all audio and visual pairs, even though many pairs are irrelevant, as shown by the gray dashed lines in Fig.$_{\!}$~\ref{motivation} $_{\!}$(b). 
This becomes especially problematic in long videos where events occur briefly and sparsely, leading the model to \textit{over-attend irrelevant audio-visual pairs} while failing to capture the local temporal context around the events effectively.

The above discussions motivate us to propose a \underline{\textbf{Lo}}cality-aware cross-modal \underline{\textbf{Co}}rrespondence learning framework: \textbf{\textsc{LoCo}}, which addresses the weakness of previous attention-based DAVE methods by making full use of local cross-modal correlation in an elegant manner. The core idea is to \textit{\textbf{leverage the local temporal continuity nature of audio-visual events (i.e., local cross-modal coherence) during both unimodal and cross-modal encoding stages}} for DAVE. Specifically, close-range audio-visual segments exhibit similarity, while remote segments often remain distinct. This inherent property acts as valuable yet free supervision signals that guide the filtering of irrelevant cross-modal noise and inspire the extraction of complementary multimodal features during both unimodal and cross-modal learning stages. 

In detail, we design \textbf{Local Correspondence Feature  (LCF) Modulation} to enforce unimodal encoders to focus on modality-shared semantics by maximizing agreement between audio and visual features. 
LCF leverages local cross-modal coherence and imposes unequal attraction, pulling positive pairs (\ie, cross-modal features in the same video) with stronger similarity more tightly while relaxing constraints on less relevant ones, thus promoting more precise cross-modal alignment.
To better aggregate such semantically aligned audio and visual features from unimodal encoders, we further introduce \textbf{Local Adaptive Cross-modal (LAC) Interaction}. 
LAC adaptively aggregates event-specific cross-modal features via adaptive window-based attention mechanism. Rather than relying on global or predefined attention regions~\cite{zhang2022actionformer}, LAC adjusts attention regions by Window Adaptation module in a data-driven manner, enhancing intra-event coherence and accommodating events of different durations in the long video.

By incorporating LCF and LAC, \textsc{LoCo} automatically mines event-valuable information and filters out irrelevant cross-modal noise to help precise detection with the guidance of local cross-modal coherence. We evaluate our \textsc{LoCo} on DAVE benchmark UnAV-100~\cite{geng2023dense} and AVEL datasets AVE~\cite{AVE} and VGGSound-AVEL100k~\cite{zhou2022contrastive}. Experiments prove that \textsc{LoCo} surpasses state-of-the-art competitors across different metrics, \eg, \textbf{4.3\%} mAP@0.9 gains on ONE-PEACE backbone~\cite{onepeace} and \textbf{2.2\%} mAP@0.5 gains on I3D-VGGish backbone~\cite{i3d, vggish} on UnAV-100~\cite{geng2023dense}. 
Furthermore, the visualization of our localization results demonstrates that, compared to the baseline, our method more effectively filters out interference from single-modal and background events, achieving more precise event localization.

Overall, our$_{\!}$ contributions$_{\!}$ are$_{\!}$ summarized $_{\!}$as $_{\!}$follows:
\begin{itemize}
    \item We leverage local cross-modal coherence for DAVE, which serves as informative yet free supervision signals to guide the extraction of event-related information from multimodal inputs during both unimodal and cross-modal encoding stages. 
    \item The proposed Local Correspondence Feature Modulation enables unimodal encoders to capture shared cross-modal semantics by leveraging local audio-visual correlations without requiring any manual labels. 
    \item We devise Local Adaptive Cross-modal Interaction to adaptively aggregate event-related cross-modal features in a data-driven manner, which strengthens the grasp of local continuity patterns in audio-visual events.
\end{itemize}
\section{Related Work}
\label{related work}
\subsection{Audio-Visual Event Localization} 
Audio-Visual Event Localization (AVEL) is to learn a model that classifies an audible and visible event, given video and corresponding audio signals. Early AVEL approaches$_{\!}$~\cite{AVE,xu2020cross,xuan2020cross,ge2023learning} fall into the segment-level classification paradigm, highlighting action class recognition rather than precise action boundary regression. Mainstream AVEL methods can be roughly categorized into two paradigms: \textbf{i)} \textit{Single-stream paradigm}~\cite{lin2019dual, AVE, yu2022mm, zhou2021positive, liu2022dense} conduct $(C+1)$ classification at the segment level, including $C$ audio-visual event categories and one background class. \textbf{ii)} \textit{Two-stream paradigm}~\cite{wu2019dual, xia2022cross, xu2020cross, yu2021mpn, feng2023css, xue2021audio} perform $C$-class classification at the video level to identify an audio-visual event, while simultaneously carrying out binary classification at the segment level to distinguish between foreground and background. However, these methods fail to account for event-specific localization preferences, leading to unsatisfactory detection performance. To fill the gap, ~\cite{ge2023learning} introduces a new paradigm for localizing events, \ie, event-aware localization paradigm, which leverages the localization patterns of videos within the same event category to attain better localization results. Existing AVEL methods mainly concentrate on the process of audio-visual integration. 
These methods$_{\!}$~\cite{lin2019dual,wu2019dual,yu2021mpn,yu2022mm} all perform intra-modal temporal feature modeling and cross-modal feature interaction.

\subsection{Dense-localizing Audio-Visual Events} 
AVEL methods tend to identify one audio-visual event in a short trimmed video, which is unsuitable for real-world audio-visual scenes. To address the issue, $_{\!\!}$\cite{geng2023dense} proposes a \textbf{new task} (\ie, Dense-localizing Audio-Visual Events (DAVE)) and corresponding benchmark (\ie, UnAV-100). \textbf{AVEL and DAVE are designed for inherently different objectives}: AVEL aims at segment-level classification to determine whether an audio-visual event occurs within a given segment of a trimmed video. In contrast, DAVE addresses the more challenging task of segment-level regression for accurately localizing the temporal boundaries of audio-visual events in untrimmed videos. DAVE is a more challenging task with the goal of detecting multiple audio-visual events (that may co-occur and vary in length)  in a long, untrimmed video. Recent works$_{\!}$~\cite{geng2023dense,geng2024uniav}  solely rely on modality-specific encoders to first capture intra-modal temporal relation and then learn audio-visual correspondence via the dense cross-attention mechanism in a pyramid manner to obtain multi-scale discriminative audio-visual features. However, these methods model audio-visual correspondence from a global perspective and pay less attention to unimodal learning, neglecting local inductive bias, \eg, temporal prior in videos. In the image domain, existing methods~\cite{liu2021swin, liu2022swin, zhang2022vsa, qu2025learning,beltagy2020longformer,zhang2024vision} make use of spatial compactness to handle objects of different sizes in images. In contrast, our method accounts for the inherent characteristics of videos, \ie, cross-modal temporal continuity of audio-visual video sequences, so as to better capture modality-shared information during different feature representation stages. By this means, our framework boosts supervised learning of DAVE with cross-modal correspondence learning in a self-supervised and data-driven manner.  

\subsection{Uni-Modal Temporal Action Detection} 
Temporal Action Detection (TAD) aims to localize and classify all actions in an untrimmed video. Recent TAD solutions can be roughly divided into two classes: \textbf{i)} \textit{Two-stage} approaches first generate action proposals through anchor windows~\cite{buch2017sst,heilbron2016fast} or detecting action boundaries~\cite{zhao2020bottom,liu2019multi}, and then classify them into actions properly. However, they heavily rely on high-quality action proposals, hence increasing computational costs and not facilitating end-to-end training. \textbf{ii)} \textit{One-stage} approaches detect all action instances in an end-to-end manner, without using any action proposal. Recent approaches attempt to localize action instances in a DETR-like~\cite{carion2020end} style, yet dense attention in the original DETR encoder relates all segments without any inductive bias, suffering from the distribution over-smoothing problem. Thus DETR-based methods~\cite{kim2023self,tan2021relaxed,liu2019multi,shi2022react} replace standard dense attention in transformer encoder with boundary-sensitive module~\cite{tan2021relaxed}, temporal deformable attention~\cite{TadTR}, or query relation attention~\cite{shi2022react}. Apart from DETR-based solutions, another line of transformer-based works~\cite{zhang2022actionformer,shi2023tridet} learn multi-level pyramid temporal representation. Though impressive, these methods only localize visible events without the help of audio modality, neglecting both audible and visible events in real-life scenes. In contrast, our focus is to Dense-localization Audio-Visual Event (DAVE) -- a more challenging task that requires jointly addressing audio and visual information in an untrimmed video, facilitating audio-visual scene understanding. With respect to this, we capture discriminative multimodal features via exploring the local cross-modal coherence prior.


\section{Method}
\label{methodology}
\subsection{Problem Statement}
Dense-localizing audio-visual events (DAVE) aims to simultaneously identify the categories and instance boundaries (\ie, starting and ending time) for all audio-visual events, which may overlap and vary in duration within an untrimmed video. Concretely, the input is audio-visual video sequence $\mathcal{X}=\{\{A_t\},\{V_t\}\}_{t=1}^{T}$, which is represented by $T$ audio-visual segment pairs ($T$ differs among videos). $A_t$ is the audio track and $V_t$ is the visual counterpart at the $t$-th segment. The groundtruth audio-visual event set is expressed as $\!\mathcal{Y}\!=\!\{{Y}_n\!=\!(s_n,e_n,c_n)\}_{n=1}^N$, where $\!N\!$ is unique to videos. The $n$-th audio-visual event ${Y}_n$ is characterized by its starting time $s_n$, ending time $e_n$ and event label $c_n\!\in\!\{0,1\}^{C}$ ($C$ represents the number of predefined categories). Note that the constraint $s_n\!<\!e_n$ must hold. The DAVE model is expected to predict $\hat{\mathcal{Y}}\!=\!\{\hat{Y}_t\!=\!(d_t^s,d_t^e,p_t)\}_{t=1}^T$, where $p_t\!\in\!\mathbb{R}^{C}$ denotes the probabilities for $C$ event categories at time $t$, $d_t^s>0$ and $d_t^e>0$ refer to the distances from time $t$ to the start and end timestamps of the event respectively. Every timestamp $t$ in the video $\mathcal{X}$ is a potential action candidate, while $d_t^s$ and $d_t^e$ are meaningful only when an event occurs at moment $t$. The final audio-visual event localization results are calculated as follows: 
\begin{equation}
\label{eq::RESULT}
\hat{s}_t = t - d_t^s, \hat{e}_t = t + d_t^e, \hat{c}_t = \arg \max p_t.
\end{equation}

\begin{figure*}[!t]
    \centering
    \includegraphics[width=0.99\textwidth]{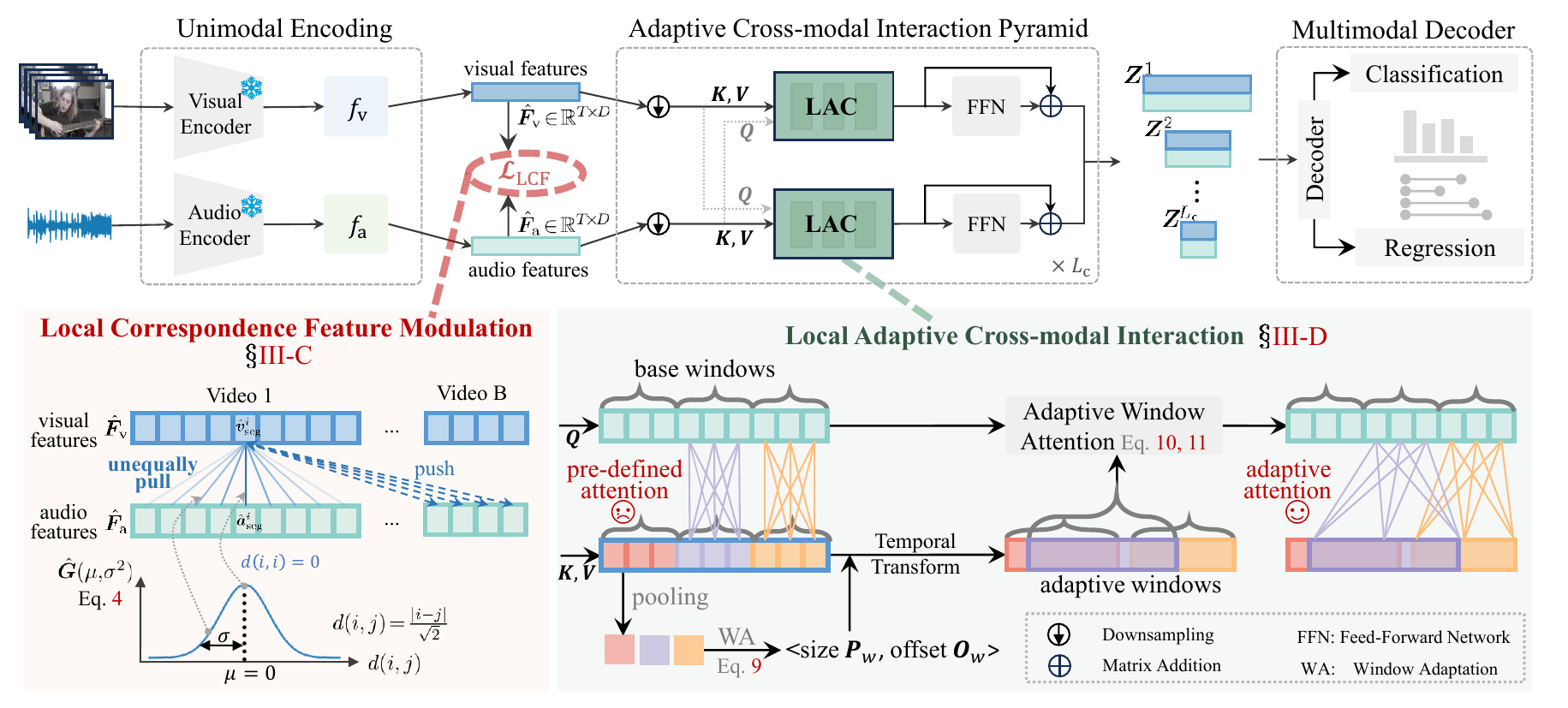} 

    \caption{\textbf{Overview of \textsc{LoCo}}. Visual and audio inputs are first processed by unimodal encoders to generate initial features. Then, \textsc{LoCo} applies LCF to pose constraints on these initial features, emphasizing modality-shared semantics. Furthermore, the adaptive cross-modal interaction pyramid adaptively adjusts cross-modal attention area based on inputs at all pyramid levels to enhance intra-event integrity, which consists of $L_\text{c}$ LAC blocks and yields multimodal feature pyramid. Finally, the multi-modal decoder identifies categories and time boundaries for audio-visual events.}
    \label{overview}
\end{figure*}

\subsection{Overall Framework} 
As illustrated in Fig.$_{\!}$~\ref{overview}, given $_{\!}$audio-visual $_{\!}$video $_{\!}$sequence $\mathcal{X}\!=\!\{\{A_t\},\{V_t\}\}_{t=1}^{T}$, our proposed $_{\!}$\textsc{LoCo} is to yield precise event localization results $\hat{\mathcal{Y}}$. Formally, $_{\!}$the $_{\!}$proposed $_{\!}$\textsc{LoCo} $_{\!}$model $_{\!}$is $_{\!}$defined by:
\begin{equation}
\label{eq::model}
\hat{{\mathcal{Y}}}=f_{\text{dec}}(f_{\text{enc}}(f_{\text{in}}(\{V_t\}_{t=1}^{T},\{A_t\}_{t=1}^{T}))),
\end{equation} 
where $f_{\text{in}}(\cdot)$ is the unimodal input encoding module, $f_{\text{enc}}(\cdot)$ refers to adaptive cross-modal interaction pyramid and $f_{\text{dec}}(\cdot)$ is multimodal decoder.

\noindent\textbf{Unimodal Input Encoding.} 
Following~\cite{geng2024uniav}, we initially employ the frozen visual and audio encoders of the pre-trained model ONE-PEACE~\cite{onepeace} to extract visual features $\bm{F}_\text{v}\!\in\!\mathbb{R}^{T\!\times\!{D}\!}$ and audio features $\bm{F}_\text{a}\!\in\!\mathbb{R}^{T\!\times\!{D}\!}$ respectively, where $D$ is the feature dimension. To capture long-term temporal relations among uni-modal segments, $\bm{F}_\text{v}$ and $\bm{F}_\text{a}$ are then fed into $L_\text{u}$ stacked unimodal transformer blocks separately, \ie, $f_{\text{v}}(\cdot)$ and $f_{\text{a}}(\cdot)$, resulting in $\hat{\bm{F}}_\text{v}\!\in\!\mathbb{R}^{T\!\times\!{D}\!}$ and $\hat{\bm{F}}_\text{a}\!\in\!\mathbb{R}^{T\!\times\!{D}\!}$. We propose Local Correspondence Feature (LCF, \cf~\S\ref{sec:LCC}) Modulation to highlight modality-shared information within an audio-visual correspondence-aware contrastive learning scheme, which poses constraints on the unimodal encoding stage.

\noindent\textbf{Adaptive Cross-modal Interaction Pyramid.} The $_{\!}$cross-modal $_{\!}$encoder $f_{\text{enc}}(\cdot)$ aggregates complementary information from $\hat{\bm{F}}_\text{v}$ and $\hat{\bm{F}}_\text{a}$ across different temporal resolutions, to address different lengths of audio-visual events. Concretely, $\hat{\bm{F}}_\text{v}$ and $\hat{\bm{F}}_\text{a}$ are processed through$_{\!}$ $L_\text{c}$ $_{\!}$Local Adaptive Cross-modal (LAC, \cf~\S\ref{SEC::CDP}) Interaction blocks with downsampling in between, producing audio-related visual feature pyramid $\mathcal{Z}_{\text{v}}\!=\!\{\bm{Z}_\text{v}^l\}_{l=1}^{L_\text{c}}$ and visual-related audio feature pyramid $\mathcal{Z}_{\text{a}}\!=\!\{\bm{Z}_\text{a}^l\}_{l=1}^{L_\text{c}}$, where $\bm{Z}_\text{v}^l$, $\bm{Z}_\text{a}^l\!\in\!\mathbb{R}^{T_l\!\times\!{D}\!}$  are outputs from $l$-th block and $T_{l-1}/T_l$ is downsampling ratio. Multimodal feature pyramid $\mathcal{Z}\!=\!\{\bm{Z}^l\}_{l=1}^{L_\text{c}}\!\in\!\mathbb{R}^{T_l\times{\text{2}D}}$ is then obtained by concatenating $\mathcal{Z}_{\text{v}}$ and $\mathcal{Z}_{\text{a}}$ at the same pyramid level. Note that, each pyramid layer is responsible for addressing events within a pre-specified time range (\eg, when the downsampling ratio is $2$, the third pyramid layer focuses on events spanning $8$ to $16$ seconds), with higher levels corresponding to longer durations. In contrast to previous methods$_{\!}$~\cite{geng2023dense,geng2024uniav} that enable dense cross-attention, LAC adaptively attends multimodal inputs to enhance intra-event integrity.

\noindent\textbf{Multimodal Decoder.} The multimodal decoder $f_{\text{dec}}(\cdot)$ generates the final detections based on multimodal feature pyramid $\mathcal{Y}\!=\!f_{\text{dec}}(\mathcal{Z})$. In our work, $f_{\text{dec}}(\cdot)$ initially conducts comprehensive fusion on $\mathcal{Z}$ at each pyramid level through transformer blocks. Classification head (\textit{Cls}) then predicts the probability of $C$ categories at each moment across all pyramid levels. Meanwhile, class-aware regression head (\textit{Reg}) calculates distances to the starting/ending time of the event at each moment for all categories, leading to regression output shape $\mathbb{R}^{2\times C \times T_l}$ at each pyramid level. As in~\cite{geng2023dense}, \textit{Cls} is implemented using three layers of 1D convolutions followed by a sigmoid function. \textit{Reg} is built with three 1D convolutions and ReLU.

\subsection{Local Correspondence Feature Modulation} 
\label{sec:LCC}
\noindent\textbf{Motivation.} Not all of the information in complex audio-visual scenarios carries equal importance$_{\!}$~\cite{duan2024cross,liu2023audio}, \eg, upon hearing a dog bark, the visual area depicting the dog should be given more focus than the region of people. Thus making full use of another modality$_{\!}$~\cite{xia2022cross, zhou2022contrastive} to guide the extraction of key information$_{\!}$ (\ie, modality-shared semantics) is helpful for further comprehending intricate audio-visual events. However, previous methods$_{\!}$~\cite{xuan2020cross,AVE,zhou2021positive, geng2023dense,geng2024uniav} separately encode visual and audio features without posing any cross-modal alignment constraint, disregarding local cross-modal coherence. With acquired visual and audio features from multimodal input encoding modules (\ie, $\hat{\bm{F}}_\text{v}, \hat{\bm{F}}_\text{a}$) in a batch, we employ Local Correspondence Feature (LCF) Modulation to pose constraints on these features, emphasizing modality-shared information. 


\noindent\textbf{Cross-modal Correspondence Feature Modulation.} Noticing the crucial role of complementary guidance from audio and visual signals in unimodal representation learning, we design Local Correspondence Feature (LCF) Modulation to maximize agreement between visual and audio features in the common space within a label-free contrastive learning scheme. Specifically, given auido-visual features $(\hat{\bm{F}}_\text{v}, \hat{\bm{F}}_\text{a}) = \{(\hat{\bm{v}}_\text{seg}^t,\hat{\bm{a}}_\text{seg}^t)\}_{t=1}^T$, let $\mathcal{B}$ denotes a batch of training video features: $\mathcal{B} = \{ (\hat{\bm{v}}_\text{seg}^i, \hat{\bm{a}}_\text{seg}^i)\}_{i=1}^{M}$, where $M\!=\!B\times T$ is the total number of segments in the batch ($B$ is batch size) and each pair $(\hat{\bm{v}}_\text{seg}^i,\hat{\bm{a}}_\text{seg}^i)$ corresponds to $\left\lceil \frac{i}{T} \right\rceil$-th segment of the $[(i-1)\!\mod\!T+1]$-th video features ($\left\lceil \cdot \right\rceil$ is ceiling function). Then, the contrastive loss function (to align visual modality with audio modality) is defined over $\mathcal{B}$ as
\begin{equation}
\label{eq:L-TSC}
    \mathcal{L}_{\text{LCF}}^{\text{ \!v2a}}\!=\!-\!\sum_{i=1}^M \sum_{j=1}^M \bm{G}_{ij}\cdot\log \frac{\exp(\langle \hat{\bm{v}}_\text{seg}^i, \hat{\bm{a}}_\text{seg}^j \rangle /\tau )}{\sum_{k=1}^M \exp (\langle \hat{\bm{v}}_\text{seg}^i, \hat{\bm{a}}_\text{seg}^k \rangle /\tau )},  
\end{equation}
where $\tau>0$ is a learnable temperature parameter, as in~\cite{li2021align,zheng2023dynamic}. $\bm{G}_{ij}$ denotes correspondence objective between $\hat{\bm{v}}_\text{seg}^i$ and $\hat{\bm{a}}_\text{seg}^j$. Before describing the calculation of $\bm{G}$, we emphasize $\bm{G}$ should ensure that values are higher for more similar pairs and $0$ for negative pairs.  By minimizing Eq.~\ref{eq:L-TSC}, audio-visual segment pairs within and across videos in the batch are considered, and positive pairs (\ie, $\bm{G}_{ij} > 0$) are attracted unequally based on their similarity degree. Note that we halve the channel dimension of features to reduce computational overhead.\\ 
\noindent\textbf{Prior-driven Correspondence Objective $\bm{G}$.} Obtaining annotations for the similarity degree of audio-visual segment pairs for untrimmed videos is almost prohibitive, $_{\!}$due $_{\!}$to $_{\!}$the $_{\!}$difficulty $_{\!}$in $_{\!}$defining $_{\!}$standardized $_{\!}$measures $_{\!}$of $_{\!}$similarity. $_{\!\!}$This $_{\!}$motivates $_{\!}$us $_{\!}$to $_{\!}$explore $_{\!}$intrinsic local cross-modal coherence$_{\!}$ within $_{\!}$a $_{\!}$video $_{\!}$(\ie, $_{\!}$cross-modal $_{\!}$segment $_{\!}$similarity $_{\!}$decays $_{\!}$as $_{\!}$the $_{\!}$segment $_{\!}$interval$_{\!}$ increases), which serves as a source of free supervision. Inspired by$_{\!}$~\cite{cao2020few,kumar2022unsupervised, nguyen2022inductive}, the cross-modal coherence within the video can be modeled by a$_{\!}$ $2$D distribution$_{\!}$ $\hat{\bm{G}}$, $_{\!}$where the marginal distribution perpendicular to the diagonal follows a Gaussian distribution centered at the intersection point on the diagonal, as
\begin{align}
\label{eq:w}
    \hat{\bm{G}}_{ij}\!\!=\!\!\frac{1}{\sigma \sqrt{2\pi}}\!\exp\!\left( -\frac{(d(i,j) - \mu)^2}{2\sigma^2} \right)\!,\!d(i,j)\!=\!\frac{|i - j|}{\sqrt{2}},
\end{align}where $\mu$ is mean parameter, $\sigma$ is standard deviation, and $d(i, j)$ measures distance between entry $(i, j)$ and diagonal line. As shown in Fig.$_{\!}$~\ref{overview}, we set $\mu\!=\!0$, ensuring synchronous audio and visual pairs are the most similar and progressively decrease perpendicular to the diagonal. A larger $\sigma$ leads to broader weights, allowing pairs that are more distant from the diagonal to still receive significant attraction. Similar to $\tau$, we set $\sigma$ as a learnable parameter, facilitating the establishment of reliable cross-modal correspondence during training. Note that we treat audio-visual segment pairs from different videos (\ie, $\left\lceil\frac{i}{T}\right\rceil \!\neq \!\left\lceil\frac{j}{T}\right\rceil$) as negative pairs, as in$_{\!}$~\cite{xia2024achieving, kimequiav, jenni2023audio}. Finally, correspondence objective $\bm{G}$ is:
\begin{equation}
\begin{aligned}
\label{eq:G-computation}
\bm{G}_{ij} &= 
\begin{cases}
\hat{\bm{G}}_{ij}, & \text{if } \left\lceil\frac{i}{T}\right\rceil = \left\lceil\frac{j}{T}\right\rceil \\
0, & \text{if } \left\lceil\frac{i}{T}\right\rceil \neq \left\lceil\frac{j}{T}\right\rceil
\end{cases}
\end{aligned}
\end{equation}

The audio-to-visual counterpart $\mathcal{L}_{\text{LCF}}^{\text{ \!a2v}}$ can be calculated in the same manner, and LCF is applied as 
\begin{equation}
\label{eq:L-TSC-all}
    \mathcal{L}_{\text{LCF}}=\frac{1}{2}(\mathcal{L}_{\text{LCF}}^{\text{ \!v2a}}+\mathcal{L}_{\text{LCF}}^{\text{ \!a2v}}).
\end{equation}  
\subsection{Local Adaptive Cross-modal Interaction}
\label{SEC::CDP}
\noindent\textbf{Core idea.} As long untrimmed videos are dominated by irrelevant backgrounds, only processing valuable segments is desirable both for speed and performance, \ie, ignores irrelevant cross-modal contents $_{\!}$~\cite{gao2020listen,he2023align}. However, previous methods learn multimodal interactions by dense cross-modal attention$_{\!}$~\cite{geng2023dense,geng2024uniav}. They ignore the local temporal continuity of audio-visual events in videos and over-attend to irrelevant audio-visual pairs, introducing semantic confusion. Thus, we devise Local Adaptive Cross-modal (LAC) Interaction in the cross-modal feature pyramid to reduce temporal redundancy in long videos. LAC learns adaptive attention areas in a data-driven manner and flexibly aggregates relevant cross-modal features.




\noindent\textbf{Base Window Construction.} In Adaptive Cross-modal Interaction Pyramid, LAC allows to better handle events of different durations at each pyramid level. As in Fig.~\ref{overview}, LAC is conducted by assigning one modality as key and value, and the other as query. We illustrate LAC with an example where audio features $\bm{Z}_\text{a}$ serve as query (visual features $\bm{Z}_\text{v}$ serve as key and value). Given $\bm{Z}_\text{v}^{l-1}$, $\bm{Z}_\text{a}^{l-1}$ (\ie, the input of $l$-th LAC block), downsampling is performed first to obtain $\bm{\tilde{Z}}_\text{v}^{l}, \bm{\tilde{Z}}_\text{a}^{l}\in \mathbb{R}^{T_l\times D}$. LAC partitions features into non-overlapping base temporal windows, \ie, $\{\bm{\tilde{Z}}_{\text{v}\_w}^{l},\bm{\tilde{Z}}_{\text{a}\_w}^{l}\!\in\!\mathbb{R}^{W\times H \times D'}\}_{w=1}^{{T_l}\!/\!{W}}$ , where $W$ is the predefined window size, $H$ is head number and $D'$ is channel dimension. Note that $D\!=\!H\!\times \!D'$. Specifically, given $\bm{\tilde{Z}}_{\text{a}\_w}^{l}, \bm{\tilde{Z}}_{\text{v}\_w}^{l}$, the query, key, and value features are got by:
\begin{equation}
    \bm{Q}_w^l = f_{\text{Linear}}(\bm{\tilde{Z}}_{\text{a}\_w}^{l}), 
\end{equation}
\begin{equation}
    \bm{K}_w^l, \bm{V}_w^l= f_{\text{Linear}}(\bm{\tilde{Z}}_{\text{v}\_w}^{l}),
\end{equation} where $\bm{Q}_w^l, \bm{K}_w^l,\bm{V}_w^l \in \mathbb{R}^{W\times H \times D'}$, and $f_{\text{Linear}}$ is Linear layer.

\noindent\textbf{Target Window Construction.} 
LAC applies Window Adaptation (WA) module to predict the ideal temporal sizes and offsets for each base window (\ie, $\bm{K}_w^l, \bm{V}_w^l$) in a data-driven manner. WA consists of average pooling, LeakyReLU~\cite{Leakyrelu} activation, and 1 x 1 convolution with stride 1 in sequence:
\begin{equation}
    \bm{P}_w, \bm{O}_w = f_{\text{convolution}}(f_{\text{LeakyReLU}}(f_{\text{average pooling}}(\bm{K}_w^l))), 
\end{equation}
where $\bm{P}_w$ and $\bm{O}_w \in \mathbb{R}^{1 \times H}$ represent the estimated temporal size and offset ($\bm{V}_w^l$ undergo the same processing). Based on $\bm{P}_w$ and $\bm{O}_w$, each base window is transformed into target window (\ie, attention area) by $H$ attention heads independently, which differs from the method on image domain~\cite{liu2021swin, liu2022swin} that window definition is shared among heads. The obtained target windows may overlap, which strengthens the ability to address overlapping events.

\noindent\textbf{Adaptive Window Attention.} Then LAC uniformly samples $W$ features from all target windows over $K^l,  V^l$ respectively. This yields $\hat{K}_{w}^{l},\hat{V}_{w}^{l}\!\in\!\mathbb{R}^{ W \times H \times D'}$ as key, value features for the query feature $Q_w^l$. The sampling count $W$ is equal to base window size, which ensures computational cost remains consistent with base window attention. To bridge connections among target windows, we adopt cross-modal sliding window attention (CSWA), the process can be defined as:
\begin{equation}
    \hat{\bm{Z}}_\text{v}^l = f_{\text{CSWA}}(Q^l, \hat{K}^{l}, \hat{V}^{l}),
\end{equation}
\begin{equation}
    \bm{Z}_\text{v}^l = \hat{\bm{Z}}_\text{v}^l + f_{\text{FFN}}(f_{\text{LN}}(\hat{\bm{Z}}_\text{v}^l)),
\end{equation}
where $Q^l, \hat{K}^{l}, \hat{V}^{l}$$\in \mathbb{R}^{T  \times  H  \times D'}$ are got by stacking $Q_w^l$, $\hat{K}_w^{l}$, $\hat{V}_w^{l}$ respectively. $f_{\text{LN}}$ is LayerNorm~\cite{ba2016layer} and $f_{\text{FFN}}$ is feed-forward network~\cite{transformer}.  

Different from recent TAD method~\cite{zhang2022actionformer} exploring the local dependency in visual modality via fix-sized hand-crafted window attention, LAC dynamically adjusts the attention area based on multimodal inputs, providing a more elegant way to process complex audio-visual scenes where events can overlap and vary in duration.



\subsection{Training and Inference}
\noindent\textbf{Loss Function.} Following~$_{\!}$\cite{geng2023dense,geng2024uniav}, we employ three losses for end-to-end optimization, \ie, focal loss~\cite{focal} for classification $\mathcal{L}_{\text{cls}}$, generalized IoU loss~\cite{generalized} for regression $\mathcal{L}_{\text{reg}}$, and Local Correspondence Feature (LCF) Modulation $\mathcal{L}_{\text{LCF}}$ ((\cf~\S\ref{sec:LCC})) . The total loss is calculated as:
\begin{equation}
\label{eq:allloss}
    \mathcal{L} = \mathcal{L}_{\text{cls}} + \mathcal{L}_{\text{reg}} + \alpha \mathcal{L}_{\text{LCF}},
\end{equation} where $\alpha$ is 0.1 by default.

\noindent\textbf{Inference.} During inference, full video sequences are fed into our model to obtain event candidates. Such event candidates are further refined by multi-class Soft-NMS~\cite{softnms} to alleviate highly overlapping temporal boundaries within the same class.
\section{Experiments}
\begin{table*}[!t]
\centering
\caption{\textbf{$_{\!}$Quantitative$_{\!}$ comparison$_{\!}$ results$_{\!}$} (see \S\ref{sec43}) on$_{\!}$ UnAV-100$_{\!}$~\cite{geng2023dense}$_{\!}$. ``ONE-PEACE" is the visual and audio encoder of ONE-PEACE~\cite{onepeace}, and ``I3D-VGGish" denotes the visual encoder is I3D~$_{\!}$\cite{i3d} and audio encoder is VGGish$_{\!}$~\cite{vggish}. The best results are bold.}
\resizebox{0.98\textwidth}{!}{
\setlength\tabcolsep{10pt}
\renewcommand\arraystretch{1.2}
\begin{tabular}{rl||c|cccccc}
\thickhline
\rowcolor{mygray} Method&& Encoder & $0.5$ & $0.6$ & $0.7$ & $0.8$ & $0.9$ & Avg. \\
\hline
\hline

VSGN~\cite{vsgn}\!\!&\!\!\!\!\!\!\!\!\!\!\!\pub{ICCV2021} & I3D-VGGish & $24.5$ & $20.2$ & $15.9$ & $11.4$ & $6.8$ & $24.1$ \\

TadTR~\cite{TadTR}\!\!&\!\!\!\!\!\!\!\!\!\!\!\pub{TIP2022} &  I3D-VGGish & $30.4$ & $27.1$ & $23.3$ & $19.4$ & $14.3$ & $29.4$ \\

ActionFormer~\cite{zhang2022actionformer}\!\!&\!\!\!\!\!\!\!\!\!\!\!\pub{ECCV2022} & I3D-VGGish& $43.5$ & $39.4$ & $33.4$ & $27.3$ & $17.9$ & $42.2$ \\

TriDet~\cite{shi2023tridet}\!\!&\!\!\!\!\!\!\!\!\!\!\!\pub{CVPR2023} &  I3D-VGGish & $46.2$ & - & - & - & - & $44.4$ \\

UnAV~\cite{geng2023dense}\!\!&\!\!\!\!\!\!\!\!\!\!\!\pub{CVPR2023} &  I3D-VGGish & $50.6$ & $45.8$ & $39.8$ & $32.4$ & $21.1$ & $47.8$ \\

UniAV(AT)~\cite{geng2024uniav}\!\!&\!\!\!\!\!\!\!\!\!\!\!\pub{arXiv2024} &  I3D-VGGish & $49.3$ & - & - & - & - & $47.0$ \\

UniAV(STF)~\cite{geng2024uniav}\!\!&\!\!\!\!\!\!\!\!\!\!\!\pub{arXiv2024} &  I3D-VGGish &  $50.1$ & - & - & - & - & $48.2$ \\

ActionFormer~\cite{zhang2022actionformer}\!\!&\!\!\!\!\!\!\!\!\!\!\!\pub{ECCV2022} &  ONE-PEACE & $49.2$ & - & - & - & - & $47.0$ \\

TriDet~\cite{shi2023tridet}\!\!&\!\!\!\!\!\!\!\!\!\!\!\pub{CVPR2023} &  ONE-PEACE & $49.7$ & - & - & - & - & $47.3$ \\

UnAV~\cite{geng2023dense}\!\!&\!\!\!\!\!\!\!\!\!\!\!\pub{CVPR2023} &  ONE-PEACE & $53.8$ & $48.7$ & $42.2$ & $33.8$ & $20.4$ & $51.0$ \\

UniAV(AT)~\cite{geng2024uniav}\!\!&\!\!\!\!\!\!\!\!\!\!\!\pub{arXiv2024} &  ONE-PEACE & $54.1$ & $48.6$ & $42.1$ & $34.3$ & $20.5$ & $50.7$ \\

UniAV(STF)~\cite{geng2024uniav}\!\!&\!\!\!\!\!\!\!\!\!\!\!\pub{arXiv2024} &  ONE-PEACE & $54.8$ & $49.4$ & $43.2$ & $35.3$ & $22.5$ & $51.7$ \\
\hline
\hline
\multicolumn{2}{c||}{\textsc{LoCo}  \textbf{(Ours)}}   &  I3D-VGGish & $\mathbf{52.8}$ & $\mathbf{47.6}$ & $\mathbf{41.1}$ & $\mathbf{33.3}$ & $\mathbf{21.9}$ & $\mathbf{49.5}$ \\

\multicolumn{2}{c||}{\textsc{LoCo}  \textbf{(Ours)}}   &  ONE-PEACE & $\mathbf{58.5}$ & $\mathbf{53.2}$ & $\mathbf{46.7}$ & $\mathbf{38.1}$ & $\mathbf{26.8}$ & $\mathbf{54.9}$ \\
\hline
\end{tabular}
}
\label{tab:sota}
\end{table*}
\subsection{Experimental Setup}
\noindent\textbf{Datasets.} UnAV-100~\cite{geng2023dense} is the only standard large-scale benchmark for DAVE task, encompassing $100$ classes across diverse domains (\eg, human activities, music, animals, vehicles, natural sounds, and tools, \etc). It contains $\!10,790\!$ videos, divided into \texttt{training}, \texttt{validation}, and \texttt{testing} sets in a $3\!:\!\!1\!:\!\!1$ ratio. Each video averages $\!2.8$ audio-visual events, annotated with categories and precise temporal boundaries. To further assess the robustness and generalization of our approach, we conduct additional evaluations on two widely used audio-visual event localization benchmarks, AVE~\cite{AVE} and VGGSound-AVEL100k~\cite{zhou2022contrastive}.

\noindent\textbf{Evaluation Metric.} For evaluation, we adopt$_{\!}$ the$_{\!}$ standard$_{\!}$ metric,$_{\!}$ \ie,$_{\!}$ mean$_{\!}$ average$_{\!}$ precision$_{\!}$ (mAP) for UnAV-100.$_{\!}$ The$_{\!}$ average mAP$_{\!}$ at$_{\!}$ temporal$_{\!}$ intersection$_{\!}$ over union (tIoU) thresholds $\left[0.1\!:\!0.1\!:\!0.9\right]$ and mAPs at tIoU thresholds $\left[0.5\!:\!0.1\!:\!0.9\right]$ are reported, as suggested by \cite{geng2023dense,geng2024uniav}. 

\begin{table}[t]

\caption{Comparison Under Both the Fully and Weakly Supervised Settings in AVE~\cite{AVE} and VGGSound-AVEL100k~\cite{zhou2022contrastive}.}
\label{table_ave}
\centering
\resizebox{0.48\textwidth}{!}{
\begin{tabular}{c|cc|cc}
\thickhline
\rowcolor{mygray}   & 
\multicolumn{2}{c|}{{AVE}} & \multicolumn{2}{c}{{VGGSound-AVEL100k}}\\
\rowcolor{mygray} \multirow{-2}{*}{Method}
 & \multicolumn{1}{c}{fully} 
& \multicolumn{1}{c|}{weakly} 
& fully 
&  weakly \\
\hline \hline
AVEL~\cite{AVE} & $68.6$       & $66.7$    & $55.7$       & $46.2$   \\
CMRA~\cite{xu2020cross} & $77.4$       & $72.9$    & $57.1$       & $46.8$   \\
MPN~\cite{yu2021mpn}  & $77.6$       & $72.0$    & -      & -  \\
PSP~\cite{zhou2021positive} & $77.8$       & $73.5$    & $58.3$       & $47.4$   \\
CPSP~\cite{zhou2022contrastive} & $78.6$       & $74.2$    & $59.9$       & $48.4$   \\
LESP~\cite{ge2023learning} & $80.4$       & $77.2$    & -     & -  \\
CACE~\cite{he2024cace} &$ 80.8$       & - & - & -      \\
\textsc{LoCo}  \textbf{(Ours)}  & $\mathbf{81.7}$& $\mathbf{79.4}$ &$\mathbf{62.1}$  & $\mathbf{50.6}$ \\
\hline
\end{tabular}
}
\end{table}

\subsection{Implementation Details}
\noindent\textbf{Network Architecture.}$_{\!}$ As$_{\!}$ with$_{\!}$ the previous method~\cite{geng2024uniav}, the sound sampling rate is $16$ kHz, and the video frame rate is $16$ FPS. The visual and audio features are extracted from the visual and audio encoders of ONE-PEACE~\cite{onepeace}, using segments of $16$ frames ($1$s)  and a stride of $4$ frames ($0.25$s). The extracted audio and visual feature dimensions are $1536$. In our model, the embedding dimension $D$ is $512$, and $L_\text{u} \!=\! 2$, $_{\!}$$L_\text{c} \!=\! 6$. $_{\!}$The $_{\!}$$_{\!}$initial $_{\!}$value $_{\!}$for $_{\!}$the $_{\!}$learnable standard $_{\!}$deviation  $_{\!}$$\sigma$ $_{\!}$is $\!1$. The head number $H=4$. $_{\!}$The $_{\!}$downsampling $_{\!}$ratio $T_{l-1}/T_l$  $_{\!}$in $_{\!}$the $_{\!}$cross-modal $_{\!}$pyramid encoder $_{\!}$is set to $\!2$, which is implemented through  a single depth-wise 1D convolution as in~\cite{geng2023dense}. Note that features at different pyramid levels correspond to detecting the audio-visual events with different time ranges. The regression head predicts distances to the starting/ending time of the audio-visual event at each moment, where the regression range is predefined for each pyramid level, following~\cite{geng2023dense,geng2024uniav}. Only if the current moment lies in an audio-visual event are the regression results valid. 

To demonstrate the adaptability of our method to different video and audio backbones, we also consider I3D~\cite{i3d} and VGGish~\cite{vggish} features, used in previous works~\cite{geng2023dense, geng2024uniav}. Identical to ~\cite{geng2023dense}, frames are sampled at $25$ FPS for each video, with the maximum length set to $224$. Then $24$ consecutive RGB frames and optical flow frames (extracted by RAFT~\cite{teed2020raft}) are input into the two-stream I3D model~\cite{i3d}, using a stride of $8$ frames. Meanwhile,  audio features are extracted using VGGish~\cite{vggish} from each $0.96$ seconds segment, employing a sliding window (stride = $0.32$ seconds) to ensure temporal alignment with the visual features.


\noindent\textbf{Training.} Consistent with previous work~\cite{geng2023dense}, we adopt the Adam optimizer~\cite{kingma2014adam} with a linear warmup of $5$ epochs. Specifically, we set the batch size to $16$, initial learning rate to $10^{-4}$ and weight decay to $10^{-4}$. To$_{\!}$ accommodate$_{\!}$ varying$_{\!}$ input$_{\!}$ video$_{\!}$ lengths, in the same way as~\cite{geng2023dense, geng2024uniav}, maximum $_{\!}$sequence$_{\!}$ length $T$ is set to a fixed value by cropping$_{\!}$ or$_{\!}$ padding, \ie, $T=256$ for ONE-PEACE~\cite{onepeace} features and $T=224$ for I3D~\cite{i3d} and VGGish~\cite{vggish} features. 

\noindent\textbf{Reproducibility.} Our model, implemented in PyTorch and python3, is trained on one RTX 3090 GPU with $24$GB memory. Testing is conducted on the same machine. 
To guarantee reproducibility, full code will be released.

\subsection{Comparison with State-of-the-Arts}
\label{sec43}
As shown in Tab.$_{\!}$~\ref{tab:sota}, \textsc{LoCo} adapts to different pre-trained models and consistently outperforms leading DAVE methods UnAV~\cite{geng2023dense} and UniAV~\cite{geng2024uniav} across all metrics on the UnAV-100~\cite{geng2023dense} dataset. Concretely, equipped with the
``ONE-PEACE" encoder, \ie, the visual and audio encoder of ONE-PEACE~\cite{onepeace}, \textsc{LoCo} yields $54.9\%$ average mAP at tIoU thresholds $\left[0.1\!:\!0.1\!:\!0.9\right]$, while the previous state-of-the-arts method, UniAV(STF)~\cite{geng2024uniav}, achieves a corresponding score of $51.7\%$. UniAV~\cite{geng2024uniav} is a unified audio-visual perception network, where UniAV(AT) denotes all-task model and UniAV(STF) refers to single-task model fine-tuned on UniAV(AT). Note that \textsc{LoCo} surpasses UniAV(STF), with \textbf{3.2\%} rise in average mAP and \textbf{4.3\%} boost in mAP@0.9 (\ie, mAP at a tight threshold of 0.9). 

Utilizing the ``I3D-VGGish" encoder (\ie, the visual encoder is I3D~\cite{i3d} and the audio encoder is VGGish~\cite{vggish}), \textsc{LoCo} still surpasses existing methods in terms of mAP at different tIoU thresholds. As seen, our method \textsc{LoCo} obtains a \textbf{2.7\%} mAP@0.5 (\ie, mAP at a threshold of 0.5) gain and a $1.3\%$  increase in average mAP, compared with UniAV(STF). Meanwhile, we compare our model \textsc{LoCo} with recent state-of-the-art TAL models, including two-stage model VSGN~\cite{vsgn} and one-stage model TadTR~\cite{TadTR}, ActionFormer~\cite{zhang2022actionformer}, and TriDet~\cite{shi2023tridet}. Consistent with~\cite{geng2023dense,geng2024uniav}, TAL methods 
are provided with concatenated audio and visual features. It can be observed that our \textsc{LoCo} outperforms all these TAL methods by a solid margin. These results demonstrate the effectiveness of our    \textsc{LoCo}. 

\begin{figure*}[t]
    \centering   \includegraphics[width=0.96\textwidth]{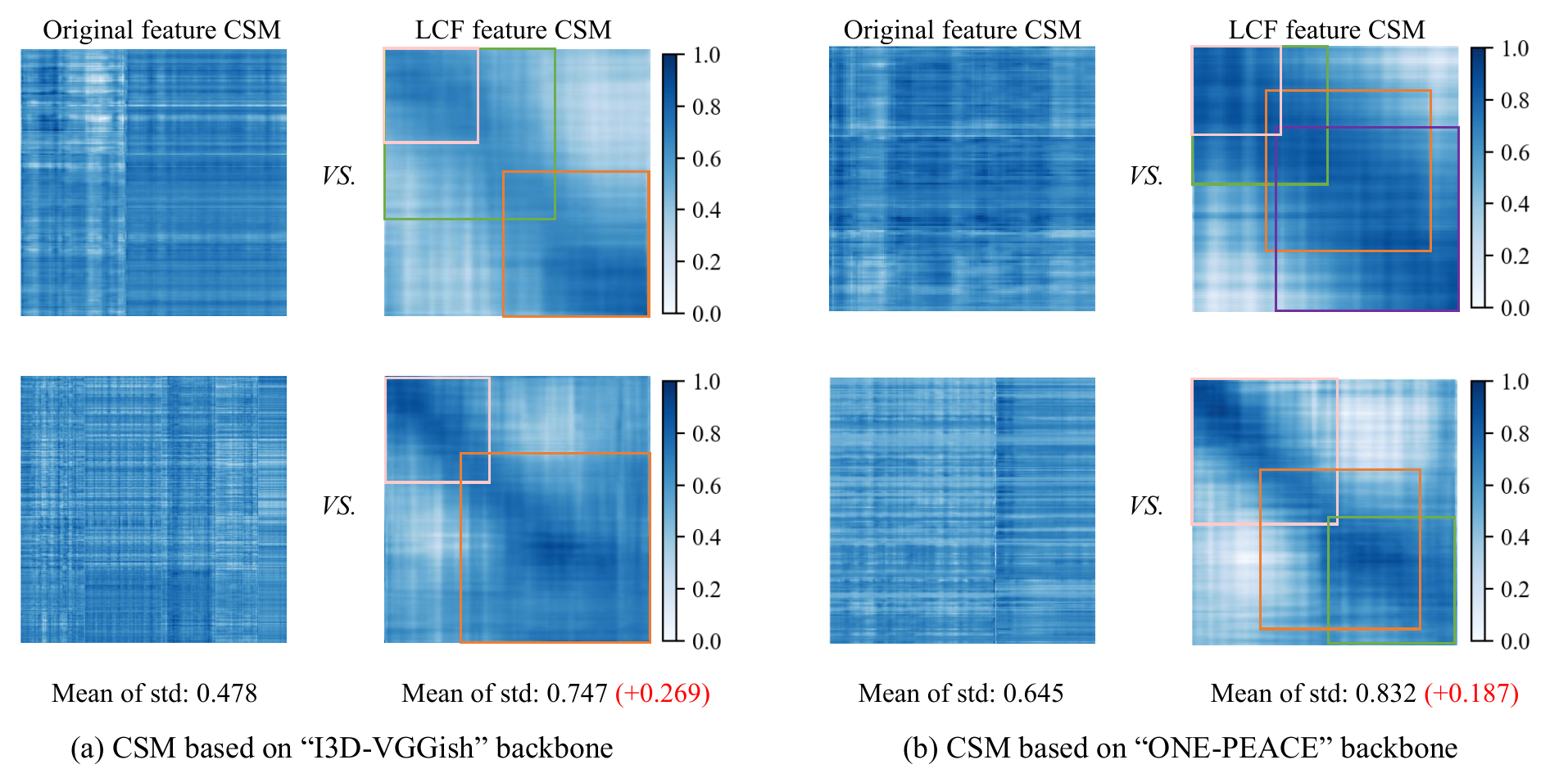} 
    
 \caption{\textbf{Qualitative results} show the effect of LCF (\cf ~\ref{sec:LCC}), which increases feature discriminability. The cross-similarity matrix (CSM) is calculated between
audio and visual features at different timestamps within the same video. For all videos in UnAV-100$_{\!}$~\cite{geng2023dense} test split, the standard deviation of the CSM is calculated, and the average of them is denoted as ``Mean of std". The increased ``Mean of std" suggests richer and more distinguishable representations. We randomly present the CSM of two videos equipped with ``I3D-VGGish" features~\cite{i3d, vggish} and ``ONE-PEACE" features~\cite{onepeace}, respectively. We also illustrate the ground-truth event boundaries using solid bounding boxes of different colors. With LCF, the audio-visual features exhibit higher cross-modal similarity within event segments, reflecting improved semantic consistency. LCF also leads to reduced similarity between audio and visual features outside the annotated event spans, promoting better discrimination between relevant and irrelevant segments.}
    \label{fig::sim_matrix}
\end{figure*}



\subsection{Evaluation on More Datasets} 

To further assess the robustness of our approach, we conduct experiments on the AVEL benchmarks, including AVE~\cite{AVE} and VGGSound-AVEL100k~\cite{zhou2022contrastive}. 
As AVEL aims at segment-level classification to determine whether an audio-visual event occurs within a segment of a trimmed video, our method uses only the classification head, without the need for the regression head. 
Given the shorter video lengths in AVE and VGGSound-AVEL100k, we set the number of pyramid levels $L_c$ to 3. 
Following CPSP~\cite{zhou2022contrastive}, we use VGG-19~\cite{simonyan2014very} and VGGish~\cite{hershey2017cnn} as visual and audio feature extractors, respectively.
Accuracy is used as the evaluation metric.
Table~\ref{table_ave} demonstrates that our method outperforms previous AVEL methods across both fully and weakly supervised settings.
This performance gain can be attributed to our dynamic perception mechanism, which effectively captures audio-visual event cues with varying temporal extents.

\subsection{Diagnostic Experiments}
\label{sec44}
\begin{figure}[!t]
    \centering   \includegraphics[width=0.3\textwidth]{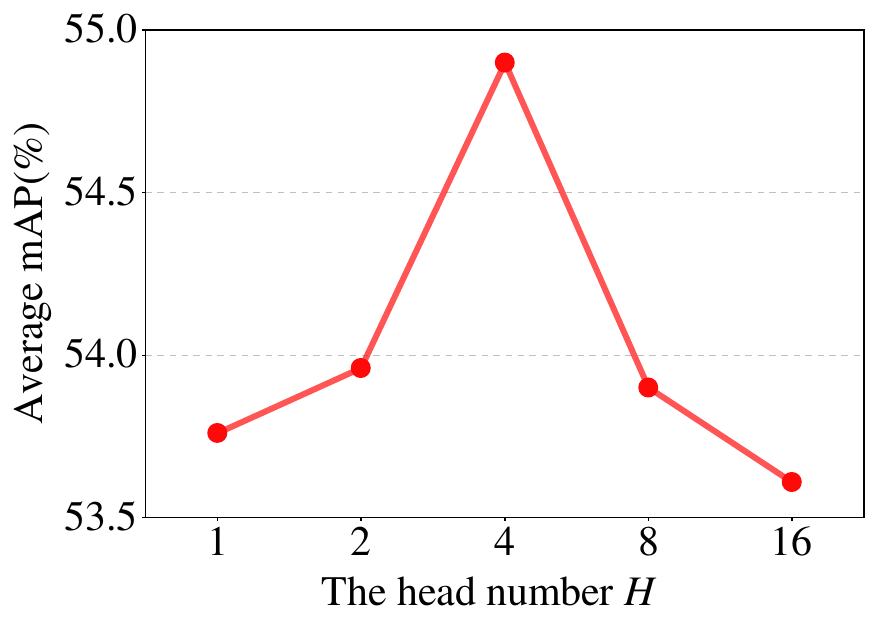} 
\caption{\textbf{The impact of head number $H$} in Local Adaptive Cross-modal (LAC) Interaction on average mAP incorporating ``ONE-PEACE" backbone~\cite{onepeace}.}
    \label{fig::n_head}
\end{figure}

\begin{figure}[!t]
    \centering   \includegraphics[width=0.3\textwidth]{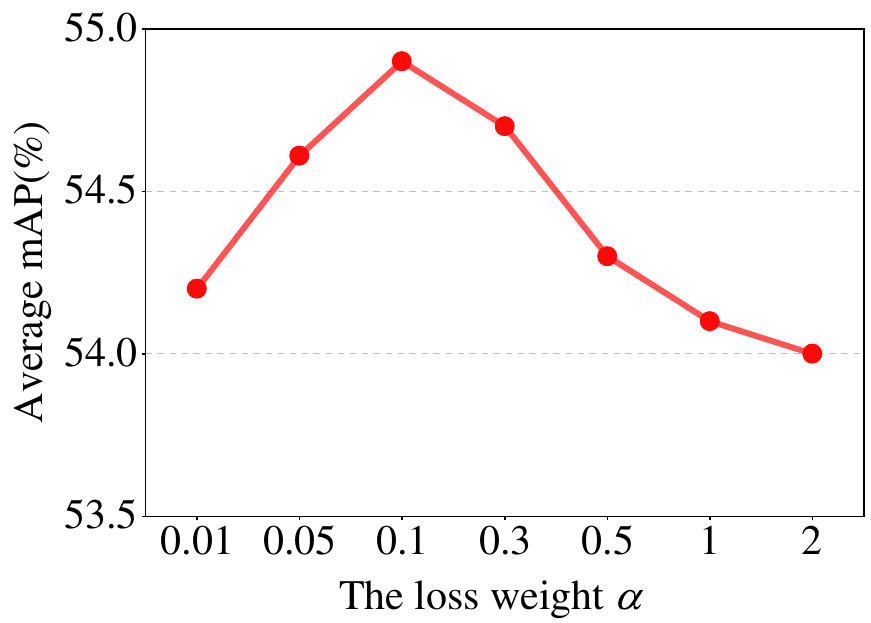} 
\caption{\textbf{The impact of parameter} $\alpha$ on average mAP built upon ``ONE-PEACE" features~\cite{onepeace}.}
    \label{loss_weight}
\end{figure}

To thoroughly evaluate our model, we conduct extensive ablation studies.
Firstly, we offer a detailed analysis of the key components of our method \textsc{LoCo}, including LCF (\cf~\S\ref{sec:LCC}) and LAC (\cf~\S\ref{SEC::CDP}), displayed in Tab.~\ref{tab:core} and Tab.~\ref{tab:core_i3dvgg}. In addition, we compare our proposed LCF and LAC with other alternatives to confirm the advantages of these components, illustrated in Tab.~\ref{tab:losstype} and Tab.~\ref{tab:wintype} respectively. We also conduct ablation experiments on key hyperparameter, \ie, the base window size $W$ in LAC, the head number $H$ in LAC, and the weight $\alpha$ in Eq.~\ref{eq:allloss}, which are reported in Tab.~\ref{tab:winsize}, Fig.~\ref{fig::n_head} and Fig.~\ref{loss_weight} separately. Finally, we compare our \textsc{LoCo} with existing state-of-the-art methods and various variants of our method regarding parameters and FLOPs, shown in Tab.~\ref{tab:floppar}.


\noindent \textbf{Key Component Analysis based on  ``ONE-PEACE" Encoder.} 
We first analyze the impact of our core designs, \ie, LCF (\cf~\S\ref{sec:LCC}) and LAC (\cf~\S\ref{SEC::CDP}) based on ``ONE-PEACE" encoder~\cite{onepeace}, which are presented in Tab.~\ref{tab:core}. The baseline in row \#1 of Tab.~\ref{tab:core} denotes our method \textsc{LoCo} without LCF and LAC. As shown in Tab.~\ref{tab:core}, additionally considering complementary guidance from audio and visual modalities (\ie, LCF) in unimodal learning stage (row \#2) leads to a substantial performance gain (\ie, $\bm{9.8}$\% mAP@0.9) compared with baseline in row \#1. Besides, our model with LCF and LAC (row \#4) outperforms baseline incorporating LAC (row \#3) by $\bm{3.3}$\% in mAP@0.9 in Tab.~\ref{tab:core}. Note that mAP@0.9 implies a stringent criterion for localization accuracy, underscoring the substantial improvements brought by LCF. The results indicate that LCF consistently improves performance, regardless of whether explicit cross-modal interactions (\ie, LAC) are incorporated. According to row \#1 and row \#3 in Tab.~\ref{tab:core}, 
LAC brings $\bm{16.6}$\% gains in mAP@0.9, highlighting the importance of the adaptive cross-attention strategy. In row \#4 of Tab.~\ref{tab:core},  \textsc{LoCo} with two core components together (\ie, LCF and LAC) achieves the best performance, confirming the joint effectiveness of them. 
\begin{table}[t]

\caption{\textbf{Ablation study on the key components} in UnAV-100~\cite{geng2023dense} with the backbone ONE-PEACE~\cite{onepeace}.}
\label{table1}
\centering
\resizebox{0.48\textwidth}{!}{
\setlength\tabcolsep{3.5pt}
\renewcommand\arraystretch{1.2}
\begin{tabular}{cc|cccccc}
\thickhline
\rowcolor{mygray}  LCF & LAC & 0.5 & 0.6 & 0.7 & 0.8 & 0.9 & Avg. \\

\hline
\hline

&  & $37.1$ & $29.0$ & $21.6$ & $13.6$ & $6.9$ & $37.0$ \\
\cmark & & $45.6$ & $38.8$ & $31.7$ & $25.3$ & $16.7$ & $45.1$ \\
 & \cmark & $57.8$ & $52.5$ & $45.1$ & $36.5$ & $23.5$ & $53.8$ \\
\cmark & \cmark & $\mathbf{58.5}$ & $\mathbf{53.2}$ & $\mathbf{46.7}$ & $\mathbf{38.1}$ & $\mathbf{26.8}$ & $\mathbf{54.9}$ \\

\hline
\end{tabular}
}

\label{tab:core}
\end{table}
\begin{table}[t]

\label{table1}
\centering
\caption{Effect of correspondence objective $\bm{G}$ in Eq.~\ref{eq:L-TSC} based on ``ONE-PEACE" features~\cite{onepeace}. (see$_{\!}$ \S\ref{sec44}).}
\resizebox{0.49\textwidth}{!}{
\setlength\tabcolsep{3pt}
\renewcommand\arraystretch{1.1}
\begin{tabular}{c|cccccc}
\thickhline
\rowcolor{mygray}  $\bm{G}$-type & 0.5 & 0.6 & 0.7 & 0.8 & 0.9 & Avg. \\

\hline
\hline

Diagonal matrix & $57.5$ & $52.2$ &$46.1$ &$37.3$ &$25.1$ &$53.5$ \\
Softened target & $57.7$  & $52.3$ & $45.3$ & $38.0$& $26.5$&$53.7$ \\
Fixed gaussian  & $58.0$ &$53.1$ &$46.5$ &$38.0$ &$26.2$ &$54.4$ \\
Adjustable gaussian & $\mathbf{58.5}$ & $\mathbf{53.2}$ & $\mathbf{46.7}$ & $\mathbf{38.1}$ & $\mathbf{26.8}$ & $\mathbf{54.9}$ \\

\hline
\end{tabular}
}

\label{tab:losstype}
\end{table}
\begin{table}[t]
\centering
\caption{ \textbf{Ablation study on the key components} in UnAV-100~\cite{geng2023dense} with the backbone ``I3D-VGGish", \ie, visual features extracted by I3D~\cite{i3d} and audio features obtained by VGGish~\cite{vggish} (See \S\ref{sec44}).}
\resizebox{0.48\textwidth}{!}{
\setlength\tabcolsep{3.5pt}
\renewcommand\arraystretch{1.2}
\begin{tabular}{cc|cccccc}
\thickhline
\rowcolor{mygray}  LCF & LAC & 0.5 & 0.6 & 0.7 & 0.8 & 0.9 & Avg. \\

\hline
\hline

&  & $34.8$ & $27.6$ & $19.7$ & $10.8$ & $3.4$ & $33.5$ \\
\cmark & & $39.8$ & $33.2$ & $27.4$ & $21.5$ & $14.4$ & $39.2$ \\
 & \cmark & $51.9$ & $46.2$ & $39.5$ & $31.0$ & $15.4$ & $48.1$ \\
\cmark & \cmark & $\mathbf{52.8}$ & $\mathbf{47.6}$ & $\mathbf{41.1}$ & $\mathbf{33.3}$ & $\mathbf{21.9}$ & $\mathbf{49.5}$ \\

\hline
\end{tabular}

}
\label{tab:core_i3dvgg}
\end{table}

\noindent \textbf{Key Component Analysis based on  ``I3D-VGGish" Encoder.} 
We also study the impact of essential components of \textsc{LoCo}, \ie, LCF (\cf~\S\ref{sec:LCC}) and LAC (\cf~\S\ref{SEC::CDP}) based on ``I3D-VGGish" encoder in Tab.~\ref{tab:core_i3dvgg}. The baseline in row \#1 of Tab.~\ref{tab:core_i3dvgg} denotes our method \textsc{LoCo} without LCF and LAC. According to Tab.~\ref{tab:core_i3dvgg}, our \textsc{LoCo} (row \#4 of Tab.~\ref{tab:core_i3dvgg})) achieves $49.5\%$ average mAP and $21.9\%$ mAP@0.9, outperforming baseline (row \#1 in Tab.~\ref{tab:core_i3dvgg})) by $16\%$ in average mAP and $18.5\%$ in mAP@0.9. By leveraging LCF to pose constraints on the unimodal encoding stage, it improves $5.0\%$ in mAP@0.5 and $11.0\%$ in mAP@0.9, as shown in rows \#1 and \#2 of Tab.~\ref{tab:core_i3dvgg}. To evaluate the effect of LAC in \textsc{LoCo} (\ie, row \#1 and row \#3 in Tab.~\ref{tab:core_i3dvgg}), it shows that LAC yields a $14.6\%$ higher average mAP than baseline. This highlights the crucial role of adaptively aggregating
event-related multimodal features. Row \#4 in Tab.~\ref{tab:core_i3dvgg}, \ie, our full model \textsc{LoCo} with LCF and LAC, obtains the best performance across all metrics, which confirms the importance of the cooperation between LCF and LAC.
All these results prove the effectiveness of our method with respect to the ``I3D-VGGish" features.

\noindent \textbf{Impact of Correspondence Objective $\bm{G}$ in Eq.~\ref{eq:L-TSC}.}
By default, we use learnable gaussian distribution (\cf~  Eq.~\ref{eq:w}), \ie, ``Adjustable gaussian" in row \#4 of Tab.~\ref{tab:losstype} to calculate $\bm{G}$, where $\sigma$ is a learnable parameter. As shown in Tab.~\ref{tab:losstype}, we evaluate three alternatives to $\bm{G}$ employing ``ONE-PEACE" features~\cite{onepeace}. \ding{182} ``Diagonal matrix" $\bm{\Lambda}$ considers only concurrent audio-visual segment pairs as positive pairs~\cite{kimequiav}, negatively impacting performance. \ding{183} ``Softened target"~\cite{pyramidclip} roughly employs label smoothing to relax the strict constraints imposed by diagonal matrix, \ie, $\bm{G}=(1-\alpha)\bm{\Lambda}+\alpha / (M-1), \alpha=0.2$. However, the equal attraction of all positive pairs hinders performance. \ding{184} ``Fixed gaussian" uses $\sigma=1$ in Eq.~\ref{eq:w} (\ie, without adjusting $\sigma$ based on input), resulting in a suboptimal solution. In terms of average mAP, \textsc{LoCo} with ``Adjustable gaussian" outperforms \textsc{LoCo} with other alternatives to $\bm{G}$ by a large margin, \eg, ``Diagonal matrix" by $1.4\%$, ``Softened target" by $1.2\%$, and ``Fixed gaussian" by $0.5\%$. We find our method surpasses all other alternatives by effectively incorporating the intrinsic, cross-modal coherence property in a data-driven manner. 

\begin{table}[t]

\centering
\caption{Effect of different attention strategies based on ``ONE-PEACE" features~\cite{onepeace} (see \S\ref{sec44}).}
\label{tab:wintype}
\resizebox{0.49\textwidth}{!}{
\setlength\tabcolsep{3pt}
\renewcommand\arraystretch{1.1}
\begin{tabular}{c|cccccc}
\thickhline
\rowcolor{mygray}  Attention strategy & 0.5 & 0.6 & 0.7 & 0.8 & 0.9 & Avg. \\

\hline
\hline
 
Global & $57.4$ & $52.2$ & $45.4$ & $37.1$ & $25.7$ & $53.8$ \\
Fixed  & $57.8$ & $52.4$ & $45.0$ & $37.2$ & $26.0$ & $53.5$ \\
Adaptive & $\mathbf{58.5}$ & $\mathbf{53.2}$ & $\mathbf{46.7}$ & $\mathbf{38.1}$ & $\mathbf{26.8}$ & $\mathbf{54.9}$ \\

\hline
\end{tabular}
}

\end{table} 
\noindent \textbf{Local Adaptive Cross-modal Interaction.} 
Tab.$_{\!}$~\ref{tab:wintype} studies the impact of Local Adaptive Cross-modal (LAC, \cf~\S\ref{SEC::CDP}) Interaction by contrasting it with vanilla cross-attention~\cite{geng2023dense} (\ie, ``Global") and fixed local cross-attention (\ie, ``Fixed"). ``Global" introduces extra noise from irrelevant backgrounds and degrades the performance compared to local attention (row \#2 - \#3 in Tab.$_{\!}$~\ref{tab:wintype}). Concretely, the average mAP of ``Global" (row \#1 in Tab.$_{\!}$~\ref{tab:wintype}) falls short of $1.1\%$ by ``Adaptive" (row \#3 in Tab.$_{\!}$~\ref{tab:wintype}), \ie, LAC. Based on our proposed LAC, we derive a variant ``Fixed" (row \#2 in Tab.$_{\!}$~\ref{tab:wintype}): only realize cross-modal sliding window attention by a fixed-size window of $8$ (the same as the base window size in LAC). As seen, our proposed LAC exhibits a \textbf{$1.4\%$} increase in average mAP relative to ``Fixed". This is because LAC offers better flexibility, allowing our model to tailor the attention based on multimodal inputs.

\noindent \textbf{Base Window Size.} 
Tab.~\ref{tab:winsize} shows the effect of base window size $W$ in LAC by increasing $W$ from 4 to 32 based on ONE-PEACE features~\cite{onepeace}. Compared with global cross-attention~\cite{geng2023dense} in row \#5 of Tab.~\ref{tab:winsize} (\ie, ``Full"), window-based attention in row \#1 - \#4 are more favored, due to high flexibility and capacity. The best results are observed with a window size of $8$. We thus set the window size $W$ to $8$ in all the experiments by default. The performance degrades when the base window size $W$ in LAC is either too large or too small, \eg, increasing $W$ from 8 to 32 leads to poorer performance (\ie, from $58.5\%$ to $57.2\%$ in mAP@0.5). This might result from the increased difficulty in adjusting the attention area when the base window size is overly large.

\begin{table}[t]
\centering
\makeatletter\def\@captype{table}\makeatother\captionsetup{font=small}
\caption{Effect of different base window size in LAC relying on ``ONE-PEACE" backbone~\cite{onepeace}.}
\resizebox{0.49\textwidth}{!}{
\setlength\tabcolsep{2.5pt}
\renewcommand\arraystretch{1.1}
\begin{tabular}{c|cccccc}
\thickhline
\rowcolor{mygray}  Base window size & 0.5 & 0.6 & 0.7 & 0.8 & 0.9 & Avg. \\

\hline
\hline
 
$4$ & $57.5$ & $52.3$ & $45.7$ & $\mathbf{38.5}$& $26.7$ & $54.3$ \\
$8$  & $\mathbf{58.5}$ & $\mathbf{53.2}$ & $\mathbf{46.7}$ & $38.1$ & $\mathbf{26.8}$ & $\mathbf{54.9}$\\
$16$ & $57.6$ & $52.6$ & $45.7$ & $38.2$ & $26.5$ & $54.2$ \\
$32$ & $57.2$ & $52.4$ & $46.1$ & $37.9$ & $26.5$ & $54.1$ \\
Full & $57.4$ & $52.2$ & $45.4$ & $37.1$ & $25.7$ & $53.8$ \\

\hline
\end{tabular}
}

\label{tab:winsize}
\end{table}
\noindent \textbf{Impact of Head Number $H$ in LAC.}
We provide an additional ablation study on the head number $H$ in Local Adaptive Cross-modal (LAC) Interaction based on ONE-PEACE backbone~\cite{onepeace}. As shown in Fig.~\ref{fig::n_head}, our model works best with $H=4$. The optimal performance at $H=4$ likely results from its alignment with the average number of events per video in UnAV-100 dataset~\cite{geng2023dense}, enabling each head to effectively model a distinct audio-visual event. Both overly large and small values of $H$ result in degraded performance. Thus, we adopt $H=4$ by default. 

\noindent \textbf{Impact of Weight $\alpha$ in Eq.$_{\!}$~\ref{eq:allloss}.} Fig. $_{\!}$~\ref{loss_weight} depicts how different $\alpha$ influences average mAP based on ONE-PEACE features~\cite{onepeace}. Average mAP rises when $\alpha$ increases and peaks at $\alpha\!=\!0.1$. Beyond this value, the average mAP declines due to the excessive weight of $\mathcal{L}_{\text{LCF}}$ relative to other loss components. Thus, we adopt $\alpha\!=\!0.1$ by default.

\begin{table}[t]

\centering
\caption{\textbf{Comparison of FLOPs and Parameters} (see \S\ref{sec44}) across different DAVE models and variants with backbone ONE-PEACE~\cite{onepeace}. ``GB" is global cross-attention~\cite{geng2023dense,geng2024uniav}. ``LAC" is Local Adaptive Cross-modal Interaction. ``LCF" is Local Correspondence Feature Modulation. }
\resizebox{0.49\textwidth}{!}{
\setlength\tabcolsep{2.5pt}
\renewcommand\arraystretch{1.2}
\begin{tabular}{c|ccc}
\thickhline
\rowcolor{mygray}  Method & FLOPs (G) &Parameters (M) & Avg. \\

\hline
\hline
UnAV	&$60.28$&	$140.79$ & $51.0$\\
UniAV(STF)   &	$32.83$&	$186.00$ & $51.7$\\ \hline
base & $18.26$ & $71.35$ & $37.0$\\
base+GB &	$31.25$ &	$102.90$ &$51.2$\\
base+LAC&	$31.25$	&$102.95$ &$53.8$\\
base+GB+LCF	&$31.45$&	$103.68$ & $53.8$\\ \hline
\textbf{Ours (base+LCF+LAC)} &$31.45$ &$103.73$  &$\mathbf{54.9}$\\

\hline
\end{tabular}
}

\label{tab:floppar}
\end{table}
\begin{figure*}[htbp]
    \centering
    \subfloat[]{%
        \includegraphics[width=0.49\textwidth]{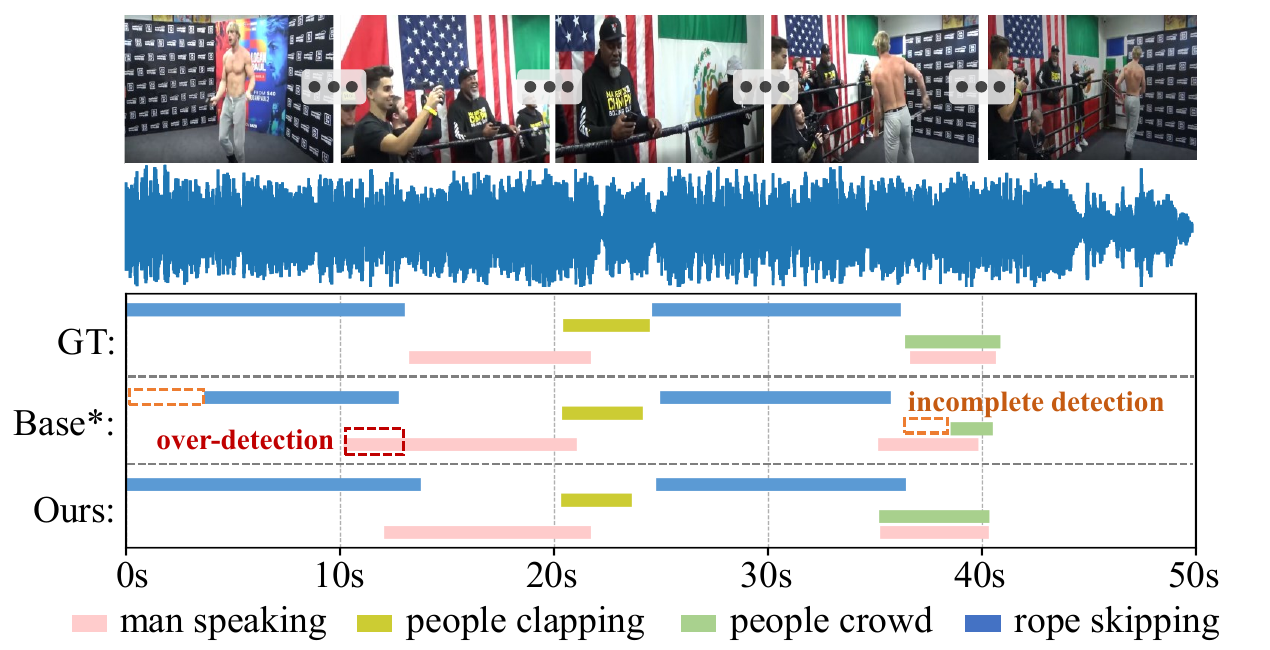}
    }
    \hfill
    \subfloat[]{%
        \includegraphics[width=0.49\textwidth]{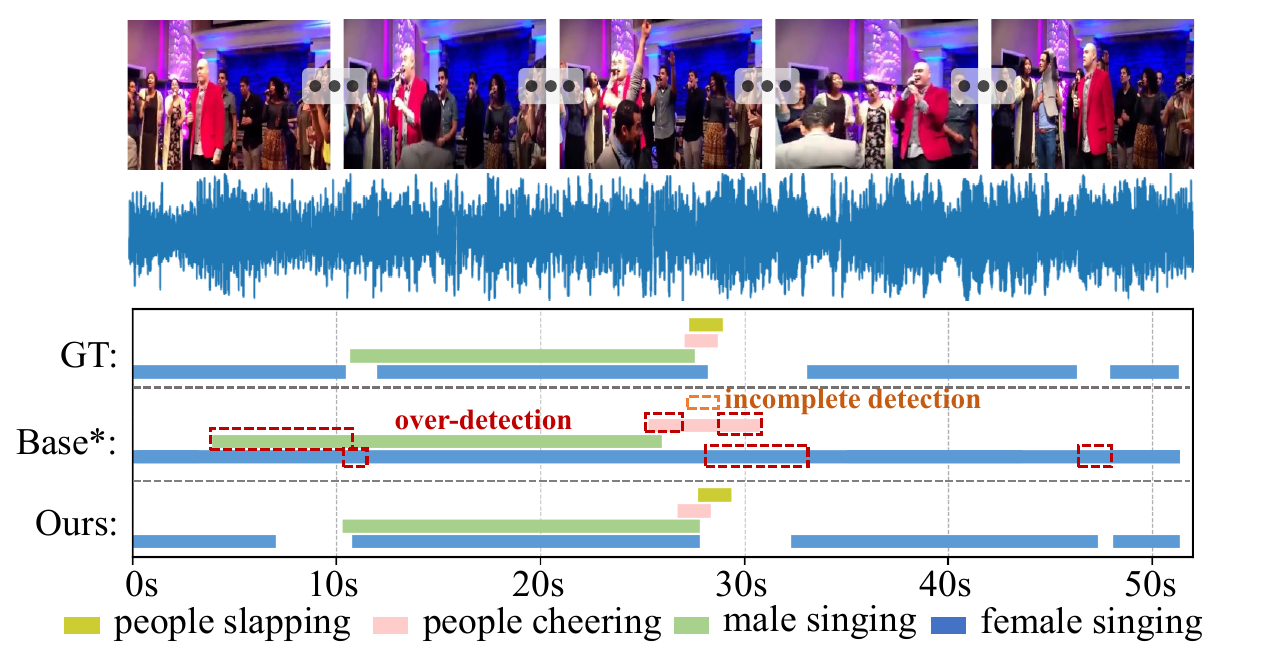}
    }


    \subfloat[]{%
        \includegraphics[width=0.49\textwidth]{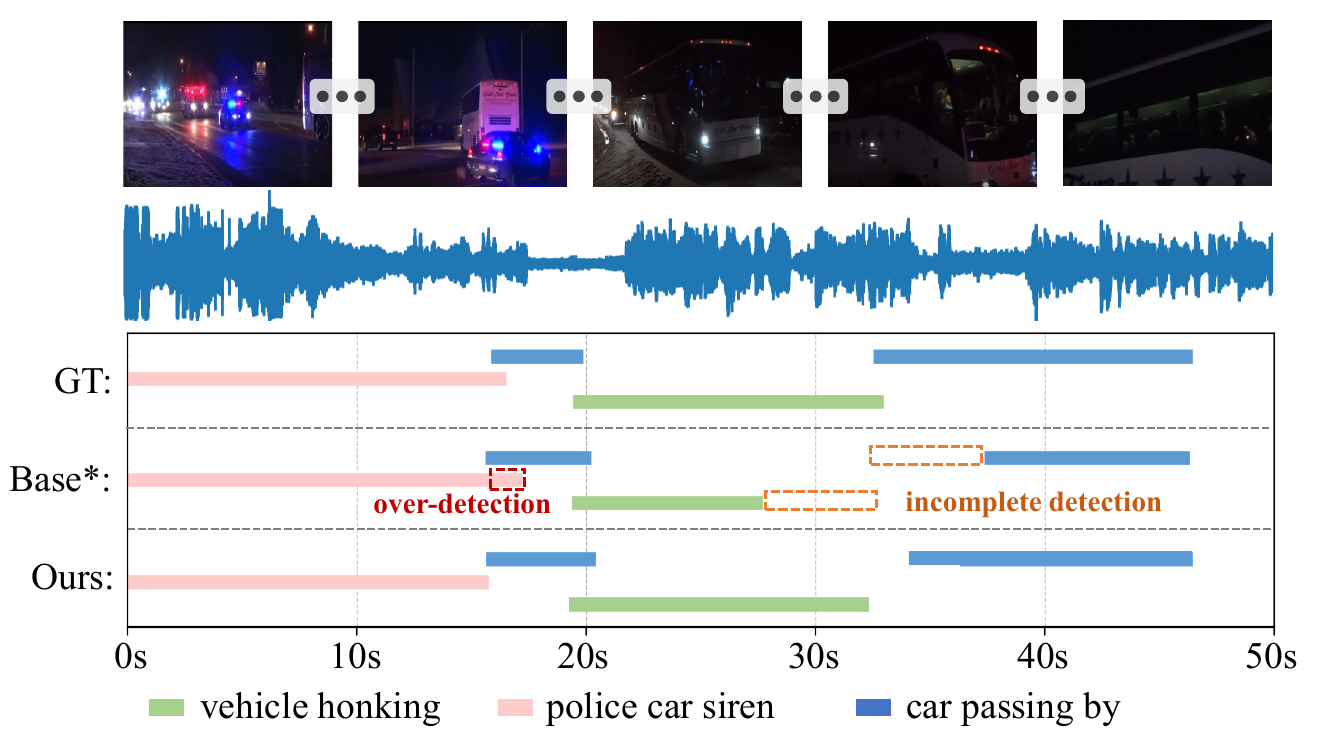}
    }
    \hfill
    \subfloat[]{%
        \includegraphics[width=0.49\textwidth]{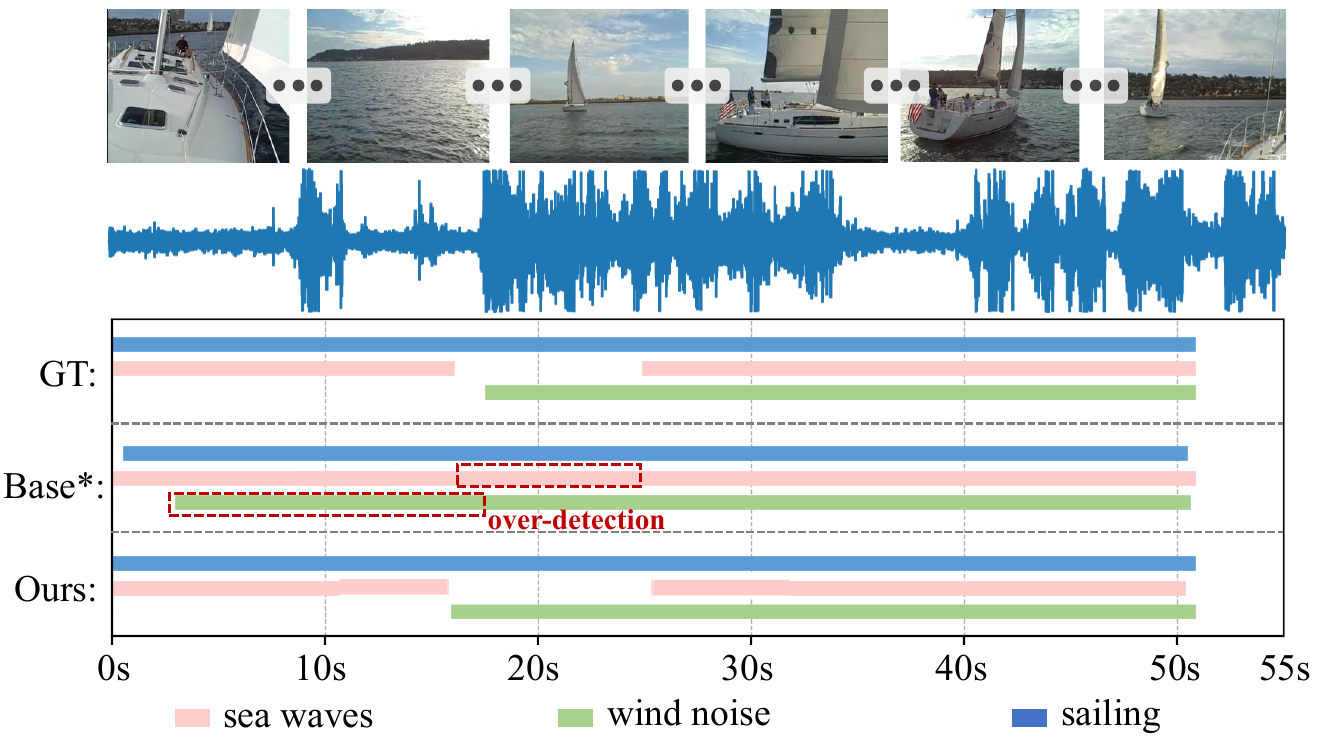}
    }

    \caption{\textbf{Qualitative detection results} on UnAV-100 test set. ``GT": ground truth. Our model displays boundaries exhibiting a high overlap with GT (See \S\ref{sec::quality}). We use red boxes to highlight the over-detected regions by Base*, and orange boxes to indicate the regions where detections are incomplete.}
    \label{fig:detection_supall}
\end{figure*}

\noindent \textbf{Parameter Analysis.}
In Tab.$_{\!}$~\ref{tab:floppar}, we compare our methods with existing state-of-the-art methods and various variants regarding parameters and FLOPs. Tab.$_{\!}$~\ref{tab:floppar} compares LAC (row \#5) with global cross attention (row \#4) used in previous methods~\cite{geng2023dense,geng2024uniav}, showing LAC slightly increased the model's parameters ($0.05$M) while bringing $2.6$\% average mAP improvement. LCF improves performance with only a minor and affordable increase in computational cost ($0.2$G FLOPs and $0.78$M parameters), as observed in rows\#5 and row \#7 in Tab.$_{\!}$~\ref{tab:floppar}. Note that compared to DAVE models (row \#1-\#2), our model has lower FLOPs and parameters, while achieving higher average mAP, \ie, realizing more precise localization results. 

\subsection{Quality Analysis} 
\label{sec::quality}
\noindent \textbf{Impact of Local Correspondence Feature Modulation.} Fig.~\ref{fig::sim_matrix} visually illustrates Local Correspondence Feature (LCF, \cf~\S\ref{sec:LCC}) Modulation enhances temporal feature discriminability by local cross-modal coherence constraint. The cross-similarity matrix (CSM) is calculated between audio and visual features from multimodal input encoding modules at different timestamps within the same video. Different from the original features (\ie, without LCF module), ``LCF features" are obtained by the model employing LCF module. We observe that LCF feature CSM exhibits a wider variety of similarities across different timestamps, demonstrating better feature discriminability. Besides, for all videos in UnAV-100$_{\!}$~\cite{geng2023dense} test split, we calculate the standard deviation of their CSM and then average them (\ie, Mean of std). We find that ``Mean of std"  increases after adopting LCF module, suggesting greater temporal sensitivity in the features~\cite{kang2023soft}. Concretely, the proposed LCF increases ``Mean of std" by $0.269$  ($0.478$ vs. $0.747$) based on ``I3D-VGGish" backbone~\cite{i3d, vggish} and raises `Mean of std" by $0.187$  ($0.645$ vs. $0.832$) based on ``ONE-PEACE" backbone, as shown in Fig.~\ref{fig::sim_matrix} (a) and (b). We illustrate the ground-truth event boundaries using solid bounding boxes of different colors in Fig.~\ref{fig::sim_matrix}. With LCF, the audio-visual features exhibit higher cross-modal similarity within event segments, reflecting improved semantic consistency. LCF also leads to reduced similarity between audio and visual features outside the annotated event spans, promoting better discrimination between relevant and irrelevant segments.

\noindent \textbf{Visualization of Localization Results.} 
 Fig.~\ref{fig:detection_supall} presents the detection results with the backbone ONE-PEACE~\cite{onepeace}. Our model achieves accurate temporal boundaries for each audio-visual event. As seen, our variant model Base* (\ie, base model equipped with global cross-modal pyramid transformer~$_{\!}$\cite{geng2023dense,geng2024uniav}) gets imprecise detection, \eg, the ``rope skipping" event in Fig.~\ref{fig:detection_supall} (a) is incompletely detected by Base*, while the ``sea waves" event in Fig.~\ref{fig:detection_supall} (d) is over-completely detected by Base*. As shown in Fig.~\ref{fig:detection_supall} (b), the ``people slapping" event is omitted, and the ``female singing" event is incorrectly localized throughout the entire video. In contrast, our model achieves more accurate temporal boundaries for each audio-visual event. This improvement is due to our model's effective extraction of modality-shared information and its deliberate suppression of background noise. 
\section{Conclusion}

In this paper, we present \textbf{\textsc{LoCo}}, a  \textbf{Lo}cality-aware cross-modal  \textbf{Co}rrespondence learning framework for Dense-localization Audio-Visual Events (DAVE). \textsc{LoCo} makes use of local cross-modal coherence to facilitate unimodal and cross-modal feature learning. The devised Local Correspondence Feature Modulation investigates $_{\!}$cross-modal $_{\!}$relations$_{\!}$ between $_{\!}$intra-and $_{\!}$inter-videos, $_{\!}$guiding $_{\!}$unimodal$_{\!}$ encoders$_{\!}$ towards modality-shared feature representation without extra annotations. To better integrate such audio and visual features, the insight from local continuity of audio-visual events in the video leads us to customize Local Adaptive Cross-modal Interaction, which adaptively aggregates event-related features in a data-driven manner. Empirical results provide strong evidence to support the effectiveness of our \textsc{LoCo}. Our work opens a new avenue for DAVE from the perspective of learning audio-visual correspondence with the guidance of local cross-modal coherence, and we wish it to pave the way for multimodal scene understanding.

\bibliographystyle{IEEEtran}
\bibliography{main}

\begin{IEEEbiography}[{\includegraphics[width=1in,height=1.25in,clip,keepaspectratio]{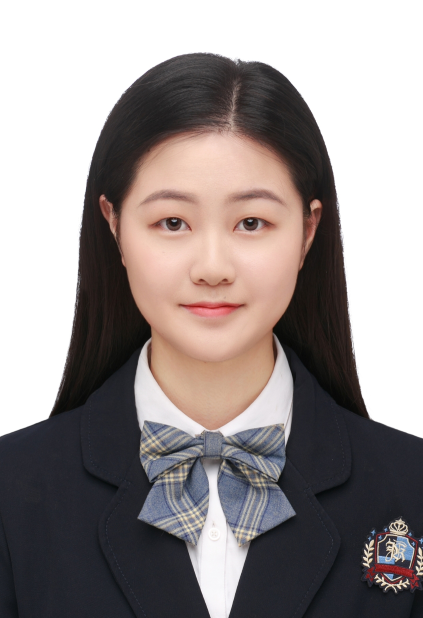}}]{Ling~Xing}
received the B.S. degree from Nanjing Forestry University, Nanjing, China. She is now a Ph.D. student in the School of Computer Science and Engineering at Nanjing University of Science and Technology. Her research interests include Video Understanding and Multimodal Learning.
\end{IEEEbiography}
\begin{IEEEbiography}[{\includegraphics[width=1in,height=1.25in,clip,keepaspectratio]{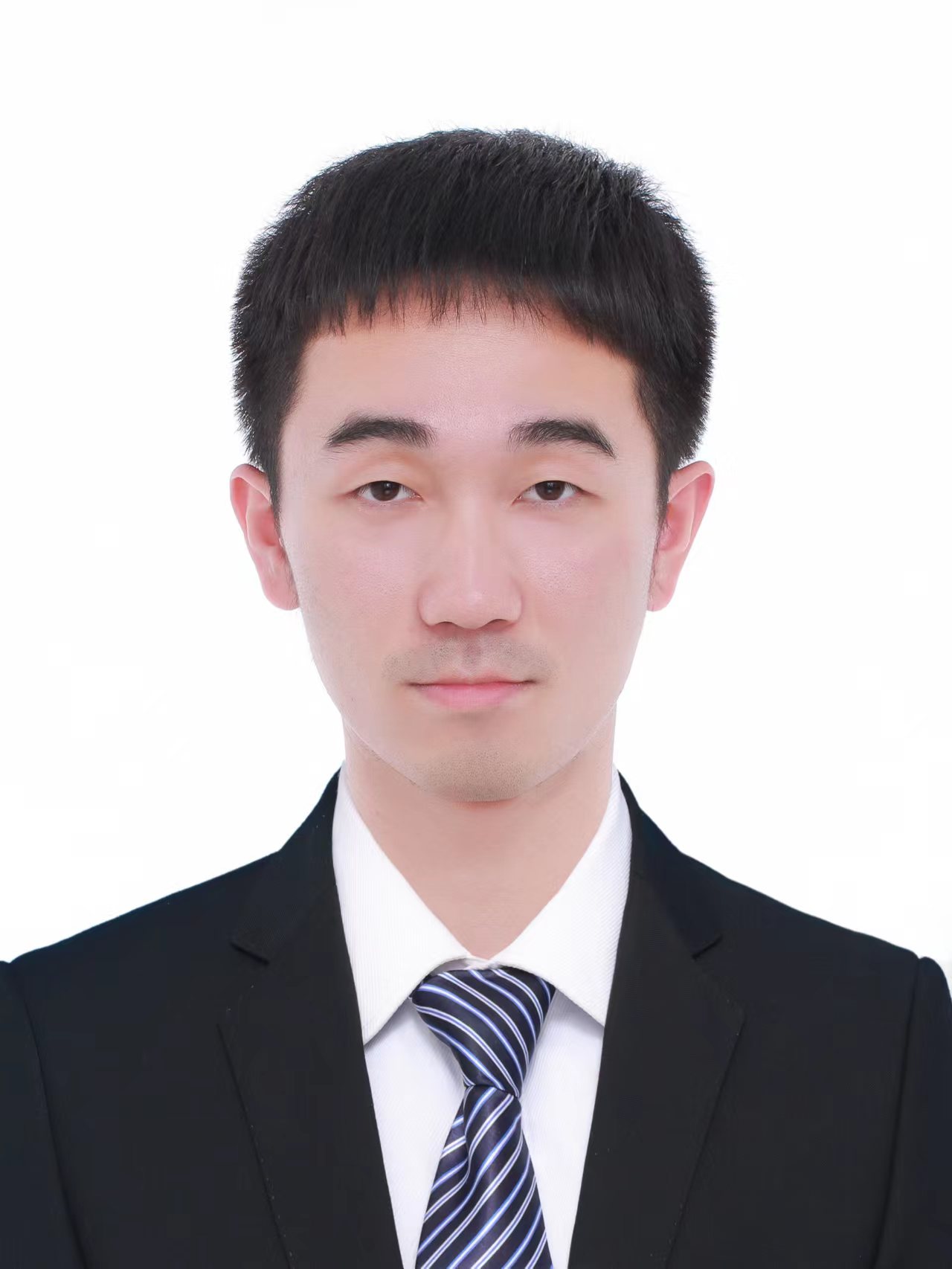}}]{Hongyu~Qu}
received the B.S. degree from Nanjing Forestry University, Nanjing, China. He is now a Ph.D. student in the School of Computer Science and Engineering at Nanjing University of Science and Technology. His research interests include Human-centric AI and Data-efficient Learning.
\end{IEEEbiography}
\begin{IEEEbiography}[{\includegraphics[width=1in,height=1.25in,clip,keepaspectratio]{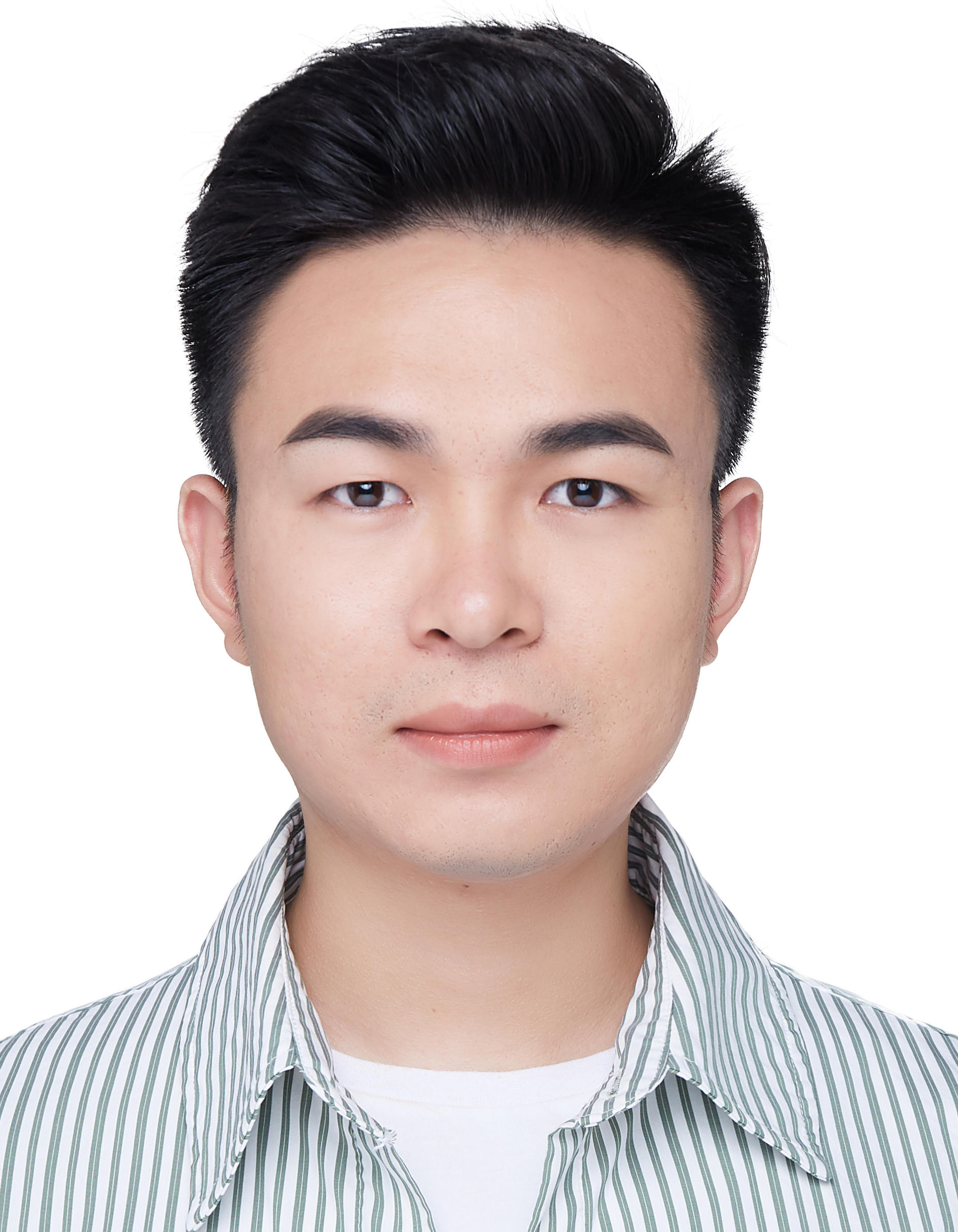}}]{Rui~Yan} 
received the Ph.D. degree at Intelligent Media Analysis Group (IMAG), Nanjing University of Science and Technology, China. He is currently an Assistant Researcher at the Department of Computer Science and Technology, Nanjing University, China. He was a research intern (part-time) at ByteDance from Jan. 2022 to Aug. 2022. He was a research intern (part-time) at Tencent from Sep. 2021 to Dec. 2021. He was a visiting researcher at the National University of Singapore (NUS) from Aug. 2021 to Aug. 2022. He was a research intern at HUAWEI NOAH'S ARK LAB from Dec. 2018 to Dec. 2019. His research mainly focuses on Complex Human Behavior Understanding and Video-Language Understanding. He has authored over 20 journal and conference papers in these areas, including IEEE TPAMI, IEEE TNNLS, IEEE TCSVT, CVPR, NeurIPS, ECCV, and ACM MM, etc.
\end{IEEEbiography}
\begin{IEEEbiography}[{\includegraphics[width=1in,height=1.25in,clip,keepaspectratio]{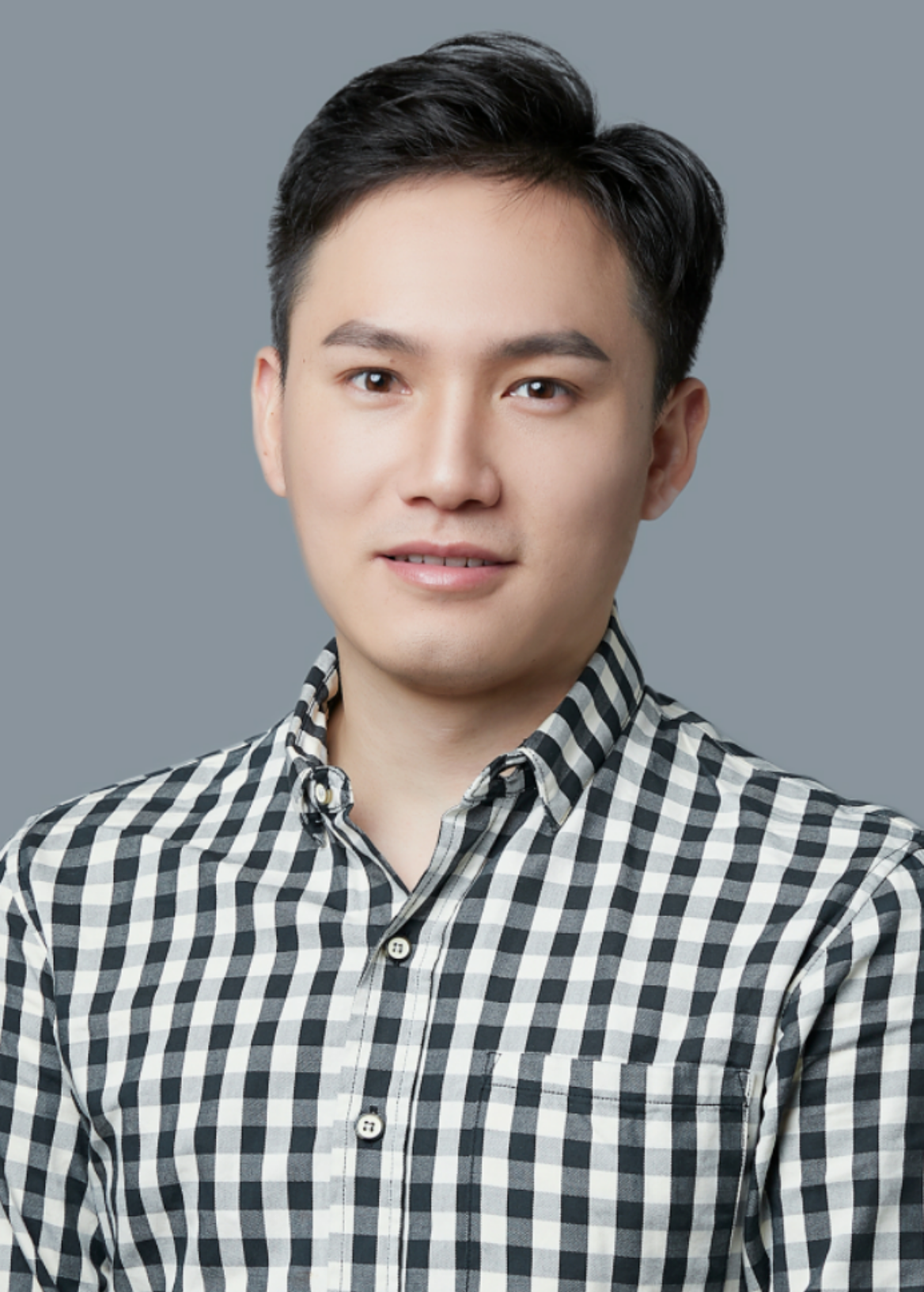}}]{Xiangbo~Shu} (Senior Member, IEEE) is currently a Professor in School of Computer Science and Engineering, Nanjing Univesity of Science and Technology, China. Before that, he also worked as a visiting scholar in National University of Singapore, Singapore. His current research interests include Computer Vision, and Multimedia. He has authored over 80 journal and conference papers in these areas, including IEEE TPAMI, IEEE TNNLS, IEEE TIP, CVPR, ICCV, ECCV, ACM MM, etc. He has received the Best Student Paper Award in MMM 2016, and the Best Paper Runner-up in ACM MM 2015. He has served as the editorial
boards of the IEEE TNNLS, and IEEE TCSVT. He is also the Member of ACM, the Senior Member of CCF, and the Senior Member of IEEE.
\end{IEEEbiography}

\begin{IEEEbiography}[{\includegraphics[width=1in,height=1.25in,clip,keepaspectratio]{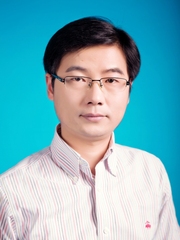}}]{Jinhui Tang}
(Senior Member, IEEE) received the
B.E. and Ph.D. degrees from the University of
Science and Technology of China, Hefei, China,
in 2003 and 2008, respectively. He is currently a
Professor with the Nanjing University of Science and
Technology, Nanjing, China. He has authored more
than 200 articles in toptier journals and conferences.
His research interests include multimedia analysis
and computer vision. Dr.Tang was a recipient of the
Best Paper Awards in ACM MM 2007 and ACM
MM Asia 2020, the Best Paper Runner-Up in ACM
MM 2015. He has served as an Associate Editor for the IEEE TNNLS, IEEE TKDE,
IEEE TMM, and IEEE TCSVT. He is a Fellow of IAPR.
\end{IEEEbiography}

\end{document}